\documentclass[sn-mathphys-num]{sn-jnl}

\usepackage{graphicx}%
\usepackage{multirow}%
\usepackage{amsmath,amssymb,amsfonts}%
\usepackage{amsthm}%
\usepackage{mathrsfs}%
\usepackage[title]{appendix}%
\usepackage{xcolor}%
\usepackage{textcomp}%
\usepackage{manyfoot}%
\usepackage{booktabs}%
\usepackage{algorithm}%
\usepackage{algorithmicx}%
\usepackage{algpseudocode}%
\usepackage{listings}%
\usepackage{subcaption} 
\usepackage{amsmath}
\usepackage{wrapfig}
\usepackage{verbatim} 
\usepackage{multicol}
\usepackage[absolute,overlay]{textpos}

\theoremstyle{thmstyleone}%
\theoremstyle{thmstyletwo}%

\theoremstyle{thmstylethree}%

\begin{document}

\title[Article Title]{Efficient Multi-robot Active SLAM}

\author*[1]{\fnm{Muhammad Farhan} \sur{Ahmed}}\email{Muhammad.Ahmed@ec-nantes.fr}
\equalcont{These authors contributed equally to this work.}

\author[2]{\fnm{Matteo} \sur{Maragliano}}\email{4636216@studenti.unige.it}
\equalcont{These authors contributed equally to this work.}

\author[1]{\fnm{Vincent} \sur{Frémont}}\email{vincent.fremont@ec-nantes.fr}

\author[2]{\fnm{Carmine Tommaso} \sur{Recchiuto}}\email{carmine.recchiuto@dibris.unige.it}

\affil*[1]{\orgdiv{Laboratoire des Sciences du Numérique de Nantes (LS2N), CNRS}, \orgname{Ecole Centrale de Nantes (ECN)}, \orgaddress{\street{1 Rue de la Noë}, \city{Nantes}, \postcode{44300}, \country{France}}}

\affil[2]{\orgname{University of Genoa}, \orgaddress{\street{5 via Balbi}, \city{Genoa}, \postcode{16126}, \country{Italy}}}

\begin{textblock*}{10cm}(5cm,1.5cm) 
\raggedleft
\footnotesize \textit{This paper has been accepted in \textbf{Journal of Intelligent and Robotic Systems (JIRS)}, kindly cite accordingly.}
\end{textblock*}

\abstract{Autonomous exploration in unknown environments remains a fundamental challenge in robotics, particularly for applications such as search and rescue, industrial inspection, and planetary exploration. Multi-robot active SLAM presents a promising solution by enabling collaborative mapping and exploration while actively reducing uncertainty. However, existing approaches often suffer from high computational costs and inefficient frontier management, making them computationally expensive for real-time applications. In this paper, we introduce an efficient multi-robot active SLAM framework that incorporates a frontier-sharing strategy to enhance robot distribution in unexplored environments. Our approach integrates a utility function that considers both pose graph uncertainty and path entropy, achieving an optimal balance between exploration coverage and computational efficiency. By filtering and prioritizing goal frontiers, our method significantly reduces computational overhead while preserving high mapping accuracy. The proposed framework has been implemented in ROS and validated through simulations and real-world experiments. Results demonstrate superior exploration performance and mapping quality compared to state-of-the-art approaches.
}




\maketitle

\section{Introduction}\label{Introduction}

Simultaneous Localization and Mapping (SLAM) involves techniques where a robot simultaneously localizes itself and maps its environment while navigating. SLAM is divided into localization, which estimates the robot's pose relative to the map, and mapping, which reconstructs the environment using information from sensors like cameras, inertial measurement units, and lidars.

Modern approaches, as described by \cite{cadena_past_2016-1}, \cite{RV20} and \cite{{RV23}} formulate the SLAM problem using a bipartite graph, where nodes represent robot or landmark poses, and edges represent measurements between them. By considering a robot with state $x \in \mathbb{R}^2$ describing its position and orientation (pose), the objective is to find the optimal state vector $\mathbf{x}^*$ which minimizes the measurement error $\mathbf
{e}_i(\mathbf{x})$ weighted by the covariance matrix $\Omega_i \in \mathbb{R}^{l \times l}$ which encapsulates the measurement uncertainty. Where $i$ is the index of edge measurement and $l$ is the dimension of the state vector as shown in Equation \ref{slam:eq1}.

\begin{equation}
\label{slam:eq1}
	\mathbf{x}^{*} = \arg \min_{\mathbf{x}} \sum_{i} \mathbf{e}_i^{T}(\mathbf{x})\Omega_i\mathbf{e}_i(\mathbf{x})
\end{equation}

Most SLAM algorithms are passive, meaning that the robot’s movement is either manually controlled and directed toward pre-defined waypoints by a human operator or external agent. In these passive systems, the navigation or path planning components do not play an active role in deciding how the robot should move to improve its mapping or localization process. The robot follows a set path or user-defined objectives without any dynamic decision-making that could optimize exploration based on the current state of the map or the robot's pose.

In contrast, Active SLAM (A-SLAM) approaches aim to address this limitation by tackling the optimal exploration problem within an unknown environment. It introduces a navigation strategy that generates future goals or target positions for the robot, with the primary objective of reducing map and pose uncertainty. Rather than passively following a fixed trajectory, the robot in A-SLAM actively selects actions that contribute to better map coverage and more accurate localization. This involves continuously assessing the current state of the map and the robot's pose, then making decisions about where the robot should move next to improve both mapping and localization.

 A-SLAM systems enable fully autonomous exploration and navigation, allowing the robot to intelligently map its environment without human intervention. This makes A-SLAM a more effective solution for autonomous robotic navigation, particularly in unknown or dynamic environments where flexibility and decision-making are crucial. In Active Collaborative SLAM (AC-SLAM) multiple robots interchange information to improve their localization estimation and map accuracy to achieve some high-level tasks such as exploration.

In this article, we extend the A-SLAM approach proposed in \cite{NP4}, to a multi-agent AC-SLAM framework and present a system for efficient environment exploration using frontiers detected over an Occupancy Grid (OG) map. In particular, in this work we aim to:
\begin{enumerate}
    \item Incorporate a utility function that leverages frontier path entropy to compute the rewards of goal candidate frontiers.
    \item Introduce an effective method for distributing robots within the environment to enhance exploration. This approach optimizes goal selection by minimizing frontiers based on reward, distance, and merged map information gain metrics.    
    \item Efficiently maximize environment exploration while maintaining a good SLAM estimates and provide a computationally inexpensive solution by reducing the number of goal frontiers.
    
\end{enumerate}
   
We implemented the proposed method in ROS, leveraging its client-server modular architecture, and made our code publicly accessible\footnote[1]{\url{https://github.com/MF-Ahmed/ACSLAM_Karto}}. Our approach is validated through both simulations and real robot experiments, demonstrating improved performances compared to state-of-the-art methods.

The article is organized as follows: Section \ref{Related Work} summarizes the related literature from selected articles. Section \ref{Methodology} presents a thorough explanation of our proposed system, with specific emphasis on the methodologies employed for frontier filtering and management, implementation of the utility function, and coordination among robots.
We show the usefulness and application of the system in simulations and real robot experiments in Sections \ref{Simulation results} and \ref{Experimental results}. Finally, in Section \ref{Conclusion} we summarize and conclude this work. Throughout this article, we will use the words \textit{robots} or \textit{agents} interchangeably, and the same applies to \textit{frontiers} and \textit{points}, as they imply the same meaning in the context.

\section{Related Work}\label{Related Work}

As previously mentioned, A-SLAM is designed for situations in which a robot must navigate in an environment that is only partially observable or unknown. In this context, the robot must choose a sequence of future actions while dealing with noisy sensor measurements that impact its understanding of both its state and the map of the environment. This scenario is typically formalized as a specific case of the Partially Observable Markov Decision Process (POMDP), as presented in \cite{thurn}, \cite{NP3}, and \cite{NP5}. 

The POMDP formulation, while widely adopted, is computationally intensive due to its consideration of planning and decision-making under uncertainty. To streamline computation, A-SLAM is usually divided into three key steps: 1) identifying potential goal positions (frontiers, i.e., boundaries between visited and unexplored areas), 2) calculating their associated costs using a utility function where utility is computed using Information Theory (IT) \cite{S21} or Theory of Optimal Experimental Design (TOED) \cite{S23}, hence selecting the next action to be performed, and 3) executing the action, eventually moving the robot to the chosen goal position.

\begin{figure}[H]
\captionsetup[subfigure]{justification=centering}
     \centering
     \begin{subfigure}[b]{0.35\textwidth}
         \centering
         \includegraphics[width=5cm,height=3.6cm]{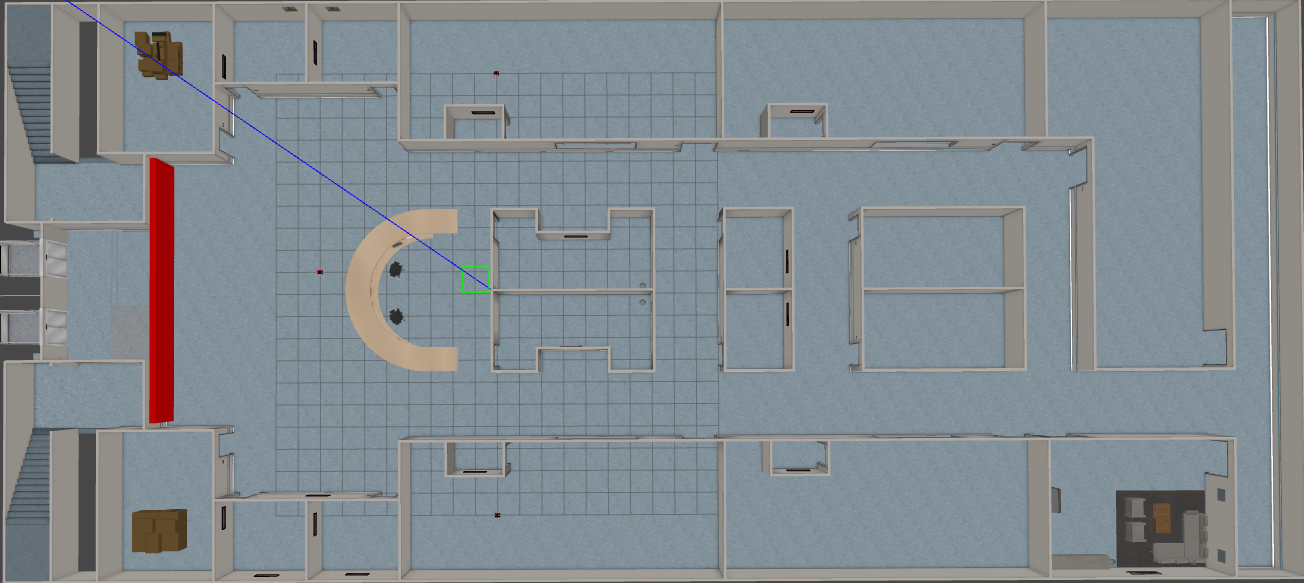}
         \caption{Simulation environment}
         \label{fig:AWS_MAP}
     \end{subfigure}
     \hfill
     \begin{subfigure}[b]{0.6\textwidth}
         \centering
         \includegraphics[width=6cm,height=3.6cm]{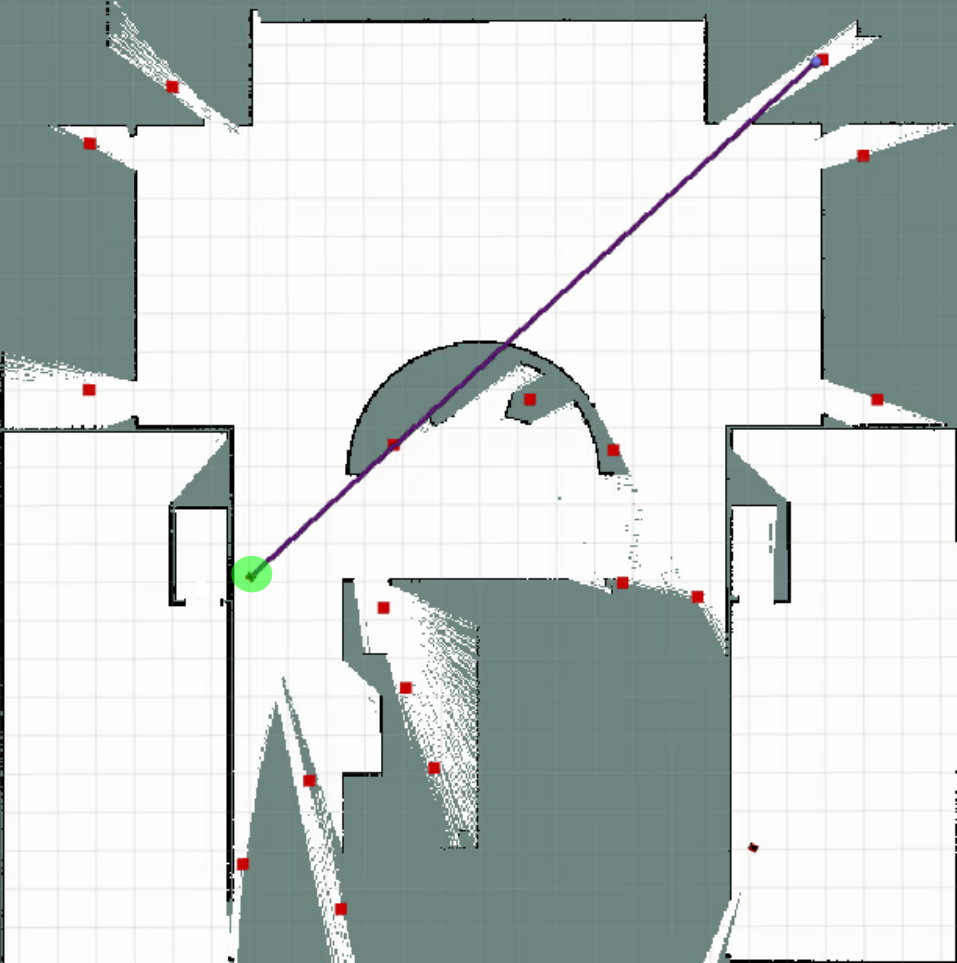}
         \caption{Computed OG map and frontiers detection}
         \label{fig:Frontiers}
     \end{subfigure}
     \hfill
        \caption{(\ref{fig:AWS_MAP}) AWS Modified Hospital environment. (\ref{fig:Frontiers}) Frontier detection on the OG map, green = robot, red = detected frontiers (centroids), white = free space, gray = unknown map area, black = obstacles}        
        \label{fig:frontierexample}
\end{figure}

Regarding the first step, a typical approach involves identifying potential exploration targets, such as frontiers. A frontier is the border between known and unknown map locations. Figure \ref{fig:frontierexample} illustrates frontier detection using lidar measurements within a simulated AWS Modified Hospital (HOS) environment\footnote[2]{\url{https://github.com/aws-robotics/aws-robomaker-hospital-world}}. 

  As the robot explores the environment, the uncertainties related to the map and robot pose increase over time, the goal is to reduce uncertainty in belief space \cite{AC21}. The existing approaches propose solutions that make use of IT and TOED approaches to quantify this uncertainty.  In IT, entropy measures the amount of uncertainty associated with a random variable or random quantity. Higher entropy leads to less information gain and vice versa. The authors in \cite{stachniss} formulate the Shannon entropy of the Map $\mathcal{M}$ as in Equation \ref{eq:6} where the map is represented as an OG and each cell $c_{i,j}$ is associated with a Bernoulli distribution $p(c_{i,j})$. The objective is to reduce the map entropy.

\begin{equation}
\label{eq:6}
	\mathcal{H}[p(\mathcal{M})] = - \sum_{i,j}(p(c_{i,j})log_2(p(c_{i,j})) + (1-p(c_{i,j}))log_2(1-p(c_{i,j}))
\end{equation}

 Alternatively, TOED defines many optimally criteria defining the mapping of the covariance matrix $\Omega_i$ to a scalar value. Hence, the priority of a set of actions is based on the amount of covariance in the joint posterior. Less covariance contributes to a higher weight of the action set. Optimality criterion deals with the minimization of average variance, reducing the volume of the covariance ellipsoid (D-optimality), and the maximum Eigenvalue as described by \cite{carrillo}. After identifying the goal positions and the utility or cost associated to reach them, the next step is to execute the optimal action that will ultimately guide the robot to the goal position. The exchanged information can be localization information \cite{AC7}, entropy \cite{stachniss}, visual features \cite{AC8}, \cite{AS4}, and frontier points \cite{AC16}.

In AC-SLAM multiple robots interchange information to improve their localization estimation and map accuracy to achieve some high-level tasks such as exploration. This collaboration raises some challenges regarding the usage of computational resources, communication resources, and the ability to recover from network failure. AC-SLAM parameters may include a) localization information \cite{AC7}, visual features \cite{AC8}, and frontier points \cite{AC16}, b) parameters relating to exploration and re-localization (to gather at a predefined meeting position) of robots as described by \cite{AC7}, c) 3D Mapping info (OctoMap) used by authors in \cite{AC16}, d) path and map entropy as used in \cite{AC17} and relative entropy, as mentioned in \cite{AC18}. Typical application scenarios include collaborative localization \cite{AC19}\cite{AC9}, exploration and exploitation strategies \cite{CS14} and trajectory planning \cite{AC17}\cite{AC18}.

The approach used by the authors in \cite{AS4} presents a multi-layer approach where the first layer selects utility criterion based on Shannon’s Entropy to goal locations (frontier points). While the second and third layers actively re-plan the path based on the updated OG map, non-linear Model Predictive Control (MPC) is applied for local path execution. In a similar approach presented by \cite{AC17}, a decentralized method for a long planning horizon of actions for exploration. The action is chosen that best minimizes the entropy change per distance traveled. At first, entropy in short horizons is computed using Square-Root Information Filter updates and that of the long horizon is computed considering a reduction in loop closures in robot paths. The main advantage of this approach is that it maintains good pose estimation and encourages loop closure trajectories.

The approach mentioned in \cite{AS12} and \cite{AS18} uses MPC to solve the area coverage and uncertainty reduction in A-SLAM. A control switching mechanism is formulated and SLAM uncertainty reduction is treated as a graph topology problem and planned as a constrained nonlinear least-squares problem. The area coverage task is solved via the sequential quadratic programming method and Linear SLAM is used for sub-map joining.

The authors in \cite{AC3} present a centralized method in which a Deep Reinforcement Learning (DRL) \cite{drl} based task allocation is used to assist agents in a relative observation task. Each agent can choose to perform its independent ORB-SLAM \cite{XP3} or help localize other agents. The reward function incorporates the influence of other agent’s transition errors in decision making. The observation function is derived from ORB-SLAM and consists of map points, keyframes, and loop closure detection components. To compute the relative observation between agents, a nonlinear optimization problem is solved using the Gauss-Newton algorithm to estimate the pose of the target agent. The large associated computational cost of this method lacks real-time application. In a similar approach by \cite{rs14112584}, the authors present a Next Best View (NBV) planner that facilitates loop closure based on information gained from visual features. The authors in \cite{AC8} propose frontiers-based coverage approaches that divide the perception task into an exploration layer and a detailed mapping layer, making use of heterogeneous robots to carry out the two tasks and solving a Fixed Start Open Traveling Salesman Problem (FSOTSP). In \cite{AC16} the authors propose frontier-based viewpoints as a valid solution to build a volumetric model of the environment with multiple agents.
  
 In a less computationally expensive approach, the method proposed by \cite{AC14} uses a Convolutional Neural Network (CNN) for fusing the aerial and ground robots maps in a traversability mapping scenario. It formulates an active perception module that uses conditional entropy to guide the robots toward high entropy paths.

In an interesting approach, the method presented in \cite{AC19} uses multiple humanoid robots, where each robot has two working modes, independent and collaborative. Each robot has two threads running simultaneously: a) the motion thread and b) the listening thread. During the motion thread, it will navigate the environment via the trajectory computed by the organizer (central server) using a D* path planner and a control strategy based on DRL and a greedy algorithm. During the listening thread, it will receive its updated pose from the organizer and may receive the command to help other robots in the vicinity improving their localization using a chained localization method. In this method, each robot’s localization is improved by its preceding robot, and its covariance is updated depending on the measurement error between the two robots.

Recently, a new method has been developed to measure uncertainty in pose graph SLAM by using graph connectivity measures. Graph connectivity indices are computationally less expensive to measure SLAM uncertainty as compared to TOED and IT approaches discussed previously. In \cite{Khosoussi2020}, \cite{NP7}, and \cite{NP9}, the authors debate how the graphical topology of SLAM has an impact on estimation reliability. They establish a relationship between the Weighted number of Spanning Trees (WST) and the D-Optimality criterion and show that the graph Laplacian is closely related to the Fisher Information Matrix (FIM). 
The authors in \cite{NP2} and \cite{NP4} extend \cite{NP9} by debating that the maximum number of Weighted Spanning Trees (WST) is directly related to the Maximum Likelihood (ML) estimate of the underlying graph SLAM problem. Instead of computing the D-optimality criterion defined over the entire slam sparse information matrix, it is computed over the weighted graph Laplacian, and it is proven that the maximum number of WST of this weighted graph Laplacian is directly related to the underlying pose graph uncertainty.

In \cite{AC9}, the authors utilize the graph connectivity indexes and propose a method for identifying weak connections in pose graphs to enhance information exchange when robots are in proximity. The proposed system identifies the weak connections in the target robot pose graph, and when the covariance increases to a certain threshold, other agents help to rectify these weak connections and generate trajectories using Rapidly Exploring Random Trees (RRT) to decrease uncertainty and improve localization. This method uses continuous refinement along with the D-optimality criterion to collaboratively plan trajectories. A bidding strategy is defined, which selects the winning host robot based on the least computational cost, feasible trajectory, and resource-friendly criteria.

 The approaches discussed above exhibit several significant limitations that warrant attention. First, these methods often incur high computational costs associated with processing a large number of frontiers. As the number of frontiers increases, the computational burden can escalate, leading to higher processing times and reduced efficiency. Second, many of these approaches fail to effectively promote the distribution of robots throughout the environment. This lack of effective distribution can reduce exploration tasks. Furthermore, the uncertainty in these methods is typically quantified using a scalar mapping of the entire pose graph covariance matrix. This matrix can become quite large, especially in landmark-based SLAM methods, which further exacerbates the computational cost. The size of the covariance matrix can lead to challenges in real-time processing and may limit the scalability of these approaches in more complex environments.

In addition to these issues, existing AC-SLAM methods often do not explicitly incorporate strategies for the efficient management of frontiers. This oversight can slow down map discovery and impede robot localization, ultimately reducing the overall effectiveness of the SLAM process. Addressing these limitations could lead to significant improvements in the efficiency and effectiveness of AC-SLAM strategies.

\section{Methodology}\label{Methodology}
\subsection{Overview of the proposed approach}\label{OverView of the approach}

Summarizing the AC-SLAM approaches discussed in Section \ref{Related Work}, we observe that they quantify the uncertainty using the entire map entropy and by using the full covariance matrix which renders them computationally expensive. Further, they do not favor the distribution of agents for maximum coverage requirements. For these reasons, here we propose an AC-SLAM method that encourages sparsity between agents while utilizing a utility function that takes into account the uncertain propagation not only of the pose graph (D-optimality) but also of the map (frontier path). Using our approach we manage to distribute the agents while maintaining a good SLAM estimate. Additionally, when compared to the state-of-the-art methods described in Section \ref{Related Work}, our method provides a computationally efficient solution by working on fewer frontiers, maximizing the exploration while using a utility function incorporating modern D-Optimality along with path entropy. 

\begin{figure}[H]
    \centering
    \includegraphics[width=15cm, height=7cm]{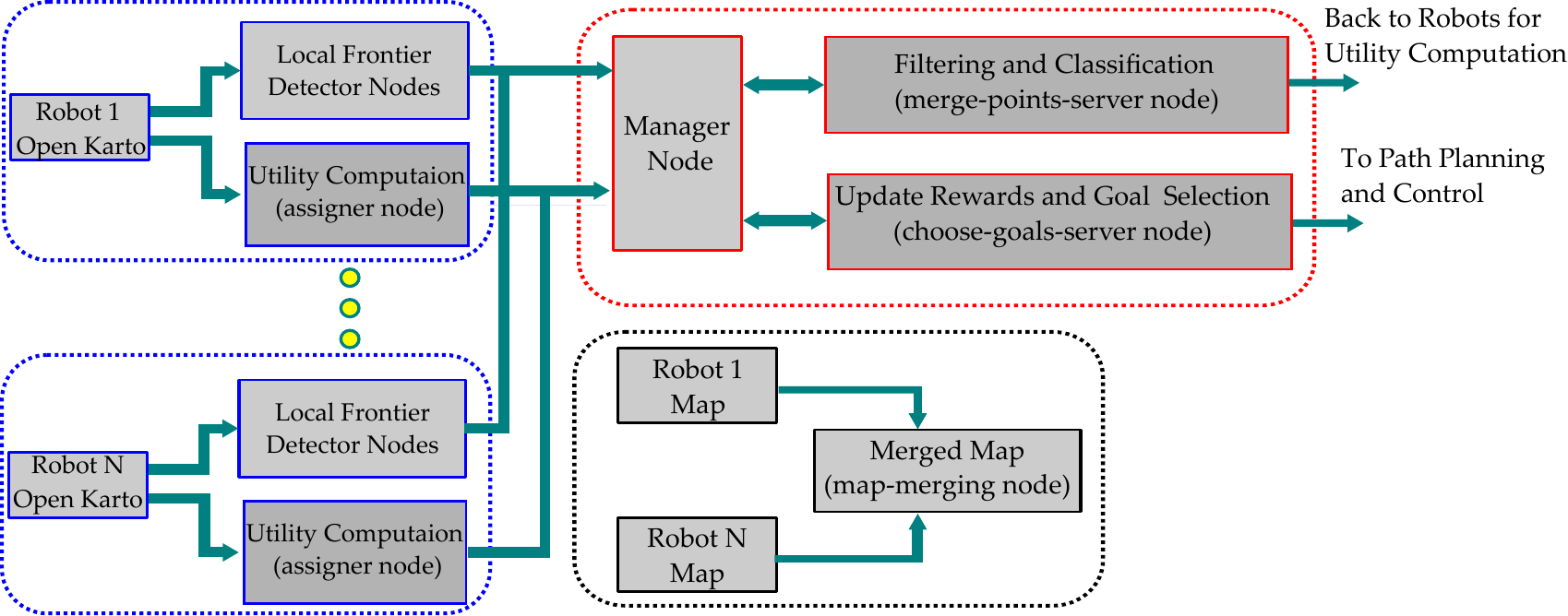}
    \caption{Architecture of the proposed method. Local nodes of each robot (blue), ROS server (red), map-merging mode (black)}
    \label{fig: flow_chart}
\end{figure}

Figure \ref{fig: flow_chart} shows the architecture and communication pipeline of our proposed approach. We have built our method upon \cite{NP4} which uses a lidar based SLAM back-end and proposes a utility function based on the modern D-Optimality criterion as a maximum number of spanning trees of the graph Laplacian of the pose graph. Each robot performs its SLAM using Open Karto \footnote[3]{\url{https://github.com/ros-perception/slam\_karto}.} and detects local frontiers relative to its map. A map merging node \footnote[4]{\url{https://github.com/robo-friends/m-explore-ros2}} (map-merging-node) merges local maps into a global map, so that all the computed frontiers are referenced to the global map. Frontiers from each agent are concatenated into a list and further processed by the Filtering and Classification (\textit{merge-points-server}) module (Section \ref{Filtering and classification}). This module removes redundant frontiers (already chosen goals from previous iteration) and further filters the frontiers/points by keeping only those points that are at (or near to) the border of the merged map (global map). This new list of points is sent back to each agent, which computes its utility and reward matrix for each point as we will see in the Utility Computation (\textit{assigner} node) module (Section \ref{Utility computation}). This reward matrix is further processed by the Update Rewards \& Goal Selection (\textit{choose-goals-server}) module (Section \ref{subsec: Update Rewards and goal selection}) which updates the rewards keeping into account the sparsity and number of already selected goal points for each agent. Finally, the selected goal for each agent is sent to the Path planning \& control module (ROS Navigation stack) which uses Dijkstra's algorithm \cite{dij} for global and Dynamic Window Approach (DWA) \cite{dwa} as local planners.   

Since the approach has been implemented in ROS and involves a centralized frontier-sharing system as shown in Figure \ref{fig: flow_chart} each agent uses two main nodes responsible for utility computation and frontier detection: the \textit{assigner} node computes the utility function and assigns the goal points to the agent, while the \textit{detector} nodes use OpenCV and RRT-based frontier detection from \cite{NP4}. The following nodes are a part of the \textit{central server} of the system: 1) the \textit{manager} node, acts as a communication gateway between robots and server, managing agent priority and frontier information, 2) the \textit{merge-points-server} node, responsible for merging lists of points acquired by different agents, and 3) the \textit{choose-goals-server} node, which chooses a specific target point from the list. By adopting a specific policy (Section \ref{subsec: Update Rewards and goal selection}), the server also aims to optimize robot distribution in the environment and minimize exploration time.

Regarding the following Sections and Algorithms, we summarize the workflow as:

\begin{enumerate}

    \item For a set of agents \( R = \{r_1, \dots, r_M\} \subset \mathbb{R}^2 \), each agent \( r_m \) detects a set of frontier points \( p_{\text{list}}^m = \{p_1^m, ..., p_N^m\} \), where each point is defined as \( p_n^m = (x_n^m, y_n^m) \in \mathbb{R}^2 \), and transmits them to a central server (subject to its availability) on a dedicated topic.

    \item The manager node takes the list of points $p_{list}$ passed by each agent and sends it to \textit{merge-points-server}.

    \item The \textit{merge-points-server} takes as input the lists received by all robots, merging the points into a unique list using Algorithm \ref{alg: list_creation}, also checking the actual frontiers on the merged map $M^{merged} = \{res, origx, origy,w,h\}$, corresponding its resolution, origin position along $x,y$ axis, width and height respectively.
    
    \item The \textit{merge-points-server} through the Algorithm \ref{alg: is_near_border} and \ref{alg: points_list_loop} limits the dimension of $p_{list}$ And eventually, it gives back the list to each agent. 
    
    \item Each agent computes the reward matrix $H$ based on the received list, using the approach described in Section \ref{subsec: Update Rewards and goal selection}, and sends it to the \textit{choose-goals-server}.

    \item The \textit{choose-goals-server} server updates the reward through Algorithms \ref{alg: select_points} and \ref{alg: update_rewards} to take into account all the points already assigned. The selected target point is fed back to the robot.  

    \item The global and local planners of the ROS package $move\_base$ are responsible for driving each agent to the selected frontier. Once the agent reaches the target, the workflow restarts from step 1.

\end{enumerate}

In the following Sections, we describe comprehensively steps 3, 4, and 5, which represent the core of the proposed approach.


\subsection{Filtering and classification}\label{Filtering and classification}

Each agent is responsible for building its map and merging local frontier points as shown in Algorithm \ref{alg: list_creation}. Some parts of the map from each agent can be overlapped and the frontiers could lie in an already mapped area when considering the merged map. Since usually the goal of an exploration task is to cover the entire area by minimizing the exploration time, the frontiers lying in the middle of the merged map are not significant because moving an agent to them would not increase the overall discovered area.  

To avoid the need to consider these points as \textit{goal-like} points, we decided to filter the points considering only the actual frontiers of the merged map. To this purpose, Algorithm \ref{alg: list_creation} takes a list of points $p_{list}$ and the merged map $M^{merged}$ and it checks if each point has enough unknown cells (\texttt{PER\_UNK}) around it, in a radius (\texttt{RAD}) (line \ref{alg: list_creation0}), i.e., if the point is near the border. If so, the point is added to the global list $uni_{pts}$. This process (line \ref{alg: list_creation0}) of Algorithm  \ref{alg: list_creation} is detailed in Algorithm \ref{alg: is_near_border}. In particular, it may be observed how the percentage \texttt{PER\_UNK} of unknown cells (Line \ref{alg: is_near_border0}) contained in the radius \texttt{RAD} is used to decide whether to keep a point in the list or not (Line \ref{alg: is_near_border1}).

\begin{figure}[!h]
\captionsetup[subfigure]{justification=centering}
     \centering
     \begin{subfigure}[b]{0.4\textwidth}
         \centering
         \includegraphics[width=5cm,height=4.0cm]{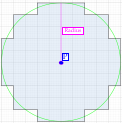}
         \caption{Discretized circle around a point \textit{P}.}
         \label{fig4:general}
     \end{subfigure}
     \hfill
     \begin{subfigure}[b]{0.45\textwidth}
         \centering
         \includegraphics[width=6.5cm,height=4cm]{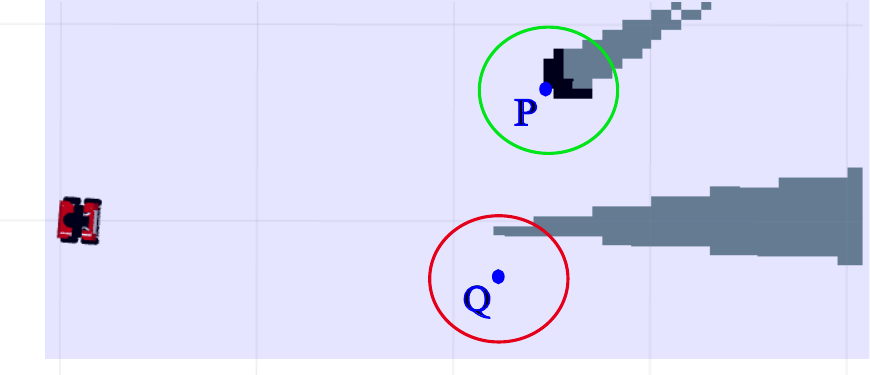}
         \caption{Discretized circle around points \textit{P} \& \textit{Q}.}
         \label{fig4:occupancy}
     \end{subfigure}
    
     \hfill
        \caption{General \ref{fig4:general} and OG Map Representation \ref{fig4:occupancy} of the Discretized circle  }
        \label{fig:ACSLAM_network}
\end{figure}

The Algorithm \ref{alg: is_near_border} starts by merging the point $p$ from the lists passed to the server. To determine if $p$ is a frontier in merged map $M^{merged}$, it first converts it into grid coordinates $(c_x, c_y)$ of $M^{merged}$ then for each cell in \texttt{RAD} it calculates the grid cell coordinates by adding its offsets to the grid coordinates of $p$ (line \ref{alg: is_near_border2}). We take a radius as shown in Figure \ref{fig4:general}, with length \texttt{RAD} around the point and check the percentage, with a value \texttt{PER\_UNK}, of unknown cells in this radius. Taking this percentage in Line \ref{alg: is_near_border0}, the algorithm can filter out or keep this point in Line \ref{alg: is_near_border1}. If this percentage $unk_{perc}$  exceeds a given threshold, the point is considered near the map's border, and the function returns "True". Otherwise, it returns "False". 

\begin{algorithm}
\small
\caption{Merge Points}
\label{alg: list_creation}

\begin{algorithmic}[1]
\Require $p_{list}$, $M^{merged}$  \Comment{$p_{list} = \{x_1,y_1,...x_N,y_N\}$, merged\_map = $M^{merged}$}
\Ensure

\State $uni_{pts} \gets 0$ 
\For{$p$ $\in$ $p_{list}$}
      \If {$N_{bdr}$$(p, M^{merged}, \texttt{RAD}, \texttt{PER\_UNK})$} \label{alg: list_creation0}
        \If{$p \notin uni_{pts}$}
            \State $uni_{pts} \gets p$ \Comment{add $p$ to unique list of points}
        \EndIf
    \EndIf
\EndFor
\end{algorithmic}
\end{algorithm}

\begin{algorithm}
\caption{Check if a Point is Near the Map Border}
\label{alg: is_near_border} 
\begin{multicols}{2}
\begin{algorithmic}[1]
\small

\Require $p, M^{merged}, \texttt{RAD}, \texttt{PER\_UNK}$: 

\Ensure $\text{Return } \textbf{True} \text{ if } p \text{ is near the border}$

\Function{$N_{bdr}$}{$p, M^{merged}, \texttt{RAD}, \texttt{PER\_UNK}$}

    \State $c_x \gets (p_x - M^{merged}_{origx}) / M^{merged}_{res}$ 
    \State $c_y \gets (p_y - M^{merged}_{origy}) / M^{merged}_{res}$ 

    \State $Rad_{c} \gets \texttt{RAD} / M^{merged}_{res}$    
    \State $Cir_{c} \gets \{\}$, $unk_{cnt} \gets 0$, $tot_{cells} \gets 0$

    \For{$i, j \in  Rad_c$}
        \State $cel_i \gets c_x + i$, $cel_j \gets c_y + j$ \label{alg: is_near_border2} 
         \If{$cel_i, cel_j \geq 0 \textbf{ \&}
        \newline \hspace*{6em} < M^{merged}_{w}, M^{merged}_{h}$}
            \State $idx \gets cel_i + cel_j \times M^{merged}_w$
            \State $Cir_{c} \gets cel_i, cel_j$
            \If{$M^{merged}_{data}[idx] = -1$}
                \State $unk_{cnt} \gets unk_{cnt} + 1$
            \EndIf
            \State $tot_{cells} \gets tot_{cells} + 1$
        \EndIf      
    \EndFor

    \If{$tot_{cells} \neq 0$}
        \State $unk_{perc} \gets \left(\frac{unk_{cnt}}{tot_{cells}}\right) \times 100$ \label{alg: is_near_border0} 
        \If{$unk_{perc} \geq \texttt{PER\_UNK}$} \label{alg: is_near_border1} 
            \State \Return \textbf{True}
        \EndIf
    \EndIf
    \State \Return \textbf{False}
\EndFunction
\end{algorithmic}
\end{multicols}
\end{algorithm}


\begin{algorithm}
\small
\caption{Check list dimension}\label{alg: points_list_loop}
\begin{multicols}{2}
\begin{algorithmic}[1]
\Require  $uni_{pts}$, $M^{merged}$, $\texttt{RAD}$, $\texttt{PER\_UNK}$
\Ensure
\While{$uni_{pts} \leq \texttt{MIN\_PTS}$ \textbf{or}  \label{alg: points_list_loop0}
 \newline \hspace*{6em} $\geq \texttt{MAX\_PTS}$}
    \If {$uni_{pts} \leq \texttt{MIN\_PTS}$}
        \State $uni_{ptsN} \gets L$,$uni_{pts} \gets \{\}$ \label{alg: points_list_loop1}
        \State $perc = perc - 10$ \label{alg: points_list_loop20}
        \For{$p \in uni_{ptsN}$}
            \If{$N_{bdr}(p,M,\texttt{RAD},\texttt{perc}$)}
             \If{$p \notin uni_{pts}$}
                    \State $uni_{pts} \gets uni_{pts} \cup \{p\}$
                \EndIf
            \EndIf
        \EndFor
    \ElsIf{$uni_{pts} \geq \texttt{MAX\_PTS}$} \label{alg: points_list_loop3}
        \State $uni_{ptsN} \gets uni_{pts}$
        \State $uni_{pts} \gets \{\}$
        \State $rad = rad + 0.25$\label{alg: points_list_loop2}
        \For{$p \in uni_{ptsN}$}
            \If{$N_{bdr}(p,M,rad, \texttt{PER\_UNK}$)}
                \If{$p \notin uni_{pts}$}
                     \State $uni_{pts} \gets uni_{pts} \cup \{p\}$
                \EndIf
            \EndIf
        \EndFor
    \EndIf
\EndWhile
\end{algorithmic}
\end{multicols}
\end{algorithm}

To illustrate this approach, Figure \ref{fig4:occupancy} presents an example featuring two points, \textit{P} and \textit{Q}, on a partially explored map. Algorithm \ref{alg: list_creation}, in conjunction with Algorithm \ref{alg: is_near_border}, identifies a circular region around each point, as shown in Figure \ref{fig4:general}, and calculates the percentage of unknown cells within the total area of the circle. Based on this computed percentage, the point is either retained or discarded according to the \texttt{PER\_UNK} threshold set during program execution. In this specific case, by appropriately configuring \texttt{PER\_UNK}, point \textit{P} is added to the global list as a border point, while point \textit{Q} is discarded.

Even discarding points that are not on the border, having many agents may lead to having extensive lists of points that need to be processed; to avoid this problem, we decided to bind the number of points to process using Algorithm \ref{alg: points_list_loop}. After the list $uni_{pts}$ is created, there is another check to validate the boundaries of the list dimension as described in Algorithm \ref{alg: points_list_loop}. If the list has fewer points than the minimum required (i.e., \texttt{MIN\_PTS}), the same is recomputed by decreasing the threshold by $10\%$ as shown in Line \ref{alg: points_list_loop1}, and \ref{alg: points_list_loop20} (where $L$ represents the length of $uni_{pts}$). Conversely, if the list has more points than a maximum threshold (line \ref{alg: points_list_loop3}) \texttt{MAX\_PTS}, it is reprocessed by increasing the radius \texttt{RAD} by 0.25m (Line \ref{alg: points_list_loop2} of Algorithm \ref{alg: points_list_loop}).

\subsection{Utility computation}\label{Utility computation}

Once the global list of frontier points is created, it is sent to the \textit{assigner} node of each agent for the computation of the utility function for each frontier candidate. We incorporate the utility function from our previous work \cite{10425063} to this multi-agent system and will briefly describe it here. It takes into account the amount of uncertainty in the map measured as path entropy and Euclidean distance to each frontier candidate. 
For each robot \( r_m \), the path entropy \( E^n_m \) for each frontier candidate \( p_n \) is computed using Equation \ref{eq:4} where \( G^n \) represents the set of grid cells in the path of the frontier candidate.

\begin{multline}\label{eq:4}
E^n_m = \textit{E}^n_m[p(c)]_{c \in G^n} = - \sum_{c \in G^n} \left( p(c_{i,j}) \log_2(p(c_{i,j})) \right. \\
\left. + p(1- c_{i,j}) \log_2(1- p(c_{i,j})), \forall c_{i,j} \in \mathcal{M} \right)
\end{multline}

We assign probability values to the OG map cells which favor exploration of unknown areas of the environment. Hence, we assign high information gain to obstacles and free space, as we are not interested in places already known to the robot.

 The path entropy is then normalized with the number of pixels/cells within the frontier path $L$. To penalize frontiers that are further away, we apply an exponential decay operator $\gamma_m^n$ as shown in Equation \ref{eq:45}, where \( d(r_m, p_n) \) is the Euclidean distance between the agent and a frontier candidate.

\begin{equation}\label{eq:45}
\gamma_m^n = \exp(-\lambda \cdot d(r_m, p_n))
\end{equation}

  Finally, the proposed utility $U_{2,m}^{n}$ as shown in Equation \ref{eq:16} is computed by weighing normalized entropy $E^n$ with $\rho^n_m$, which depends on the number of spanning trees of the weighted graph Laplacian $L_w$. Equation \ref{eq:17} represents the modern D-Optimality criterion adopted from \cite{NP4}, which proposes a utility function to quantify the best frontier candidates that have a maximum number of spanning trees in the weighted pose graph Laplacian $L_{w,m}$ towards them. Eventually, we obtain the proposed utility function $U_{m}^n$ in Equation \ref{eq:18}, which not only provides a good SLAM estimate but also increases the coverage of the unknown map by reducing the frontier path entropy.

\begin{equation}\label{eq:16}
U_{2,m}^{n} = (1-E^n/L^n)*\rho_m^n + \gamma_m^n
\end{equation}

\begin{equation}\label{eq:17}
U_{1,m}^{n} = \text{Spann}(L_{w,m})
\end{equation}

\begin{equation}\label{eq:18}
 U_{m}^n = \text{max}(U_{1,m}^{n}+U_{2,m}^{n})
\end{equation}

\begin{equation}\label{eq:19}
 H_{m} = \left[p_{list}^m, U_m^n\right]
\end{equation}

The reward matrix $H_{m}$ for each agent is computed as shown in Equation \ref{eq:19}. Where $U_m^n \subset \mathbb{R}$ is the associated reward for each frontier and is computed using Equation \ref{eq:18}.

\subsection{Update Rewards and Goal Selection}\label{subsec: Update Rewards and goal selection}

Once each agent has computed its reward matrix, 
it is sent to the \textit{choose-goals-server} (hereafter referred to as \textit{server}) to opportunely update the rewards and select the goal point of each agent. Since the server can manage one reward matrix at a time, it is asynchronously handled by the various agents, with an ascending priority: assuming to have a system of $M > 1$ agents, taking two general agents $r_1$ and $r_2$, which request the server at the same time, the goal of $r_1$ is processed before $r_2$ if $r_1 < r_2$.
Agents access the server in priority order, with the server managing one reward matrix at a time.

Moreover, the server also stores the already assigned locations, to avoid reusing them.
The server has to manage $M$ different matrices, one for each agent.
Indeed, to explore the biggest possible area of the environment, the goals for the agents need to be spread. To foster this sparsity, since there are many agents, we decided to update the reward function using Equations \ref{eq: K}-\ref{eq: R_update}, where $d$ is the Euclidean distance between the highest reward point and all other points. 

\begin{equation}
\label{eq: K}
    K = \frac{\texttt{max reward in matrix}}{\texttt{number of already chosen points}}
\end{equation}

\begin{equation}
\label{eq: k}
    k = \frac{K}{d^2}
\end{equation}

\begin{equation}
\label{eq: R_update}
    R_{new} = R_{old} - k
\end{equation}

$K$ in Equation \ref{eq: K} is motivating by two reasons:
\begin{itemize}
\item \texttt{max reward in matrix} allows for scaling $K$ with respect to the reward matrix of each agent.
\item \texttt{number of already chosen points} allows for distributing the reward update also taking into account the number of already selected points. When the number of targets already explored becomes significant, each point will only receive a smaller portion of the total reward.
\end{itemize}

It may be observed (Equation \ref{eq: R_update}) how $k$ represents a subtracting factor for the reward matrix elements, updated when a target goal for one agent is selected. Since $k$ is inversely dependent on the distance computed between the last chosen goal and the considered frontier point (Equation \ref{eq: k}), the closer the point is to the already chosen goal, the higher $k$ will be, decreasing the probability that the point will be chosen as the next goal, thus achieving the task of spreading the goals into the environment. The server then updates the reward for specific points using a negative additive factor $k$, as in Equation \ref{eq: R_update}.

The complete procedure for the selection of points (goals) is described in Algorithm \ref{alg: select_points}, also including the relative function to update rewards when a goal is selected, described in Algorithm \ref{alg: update_rewards}.

\begin{algorithm}
\small
\caption{Select Points}
\label{alg: select_points}
\begin{algorithmic}[1]

\Function{$sel_{pts}$}{$uni_{pts},uni_{ptsU}$} 
    \State $goals \gets \{\}$
    \State $H \gets$ using $uni_{pts},uni_{ptsU}$ 
    \If{$H \neq \{\}$}
        \State $p \gets$ \textbf{Point2D}() \Comment{\textbf{Point2D} is ROS message type} 
        \If{$cho_{cords} \neq \{\}$}          
            \State $H$ $\gets$ \Call{$upd\_rewards$}{$cho_{cords}$, $H$} \label{alg: select_points0} \Comment{Algorithm \ref{alg: update_rewards}} 
        \EndIf
        \State $H_{MaxID} \gets \text{get maximum reward index in }H$ 
        \State $p.x, p.y \gets \text{x,y value in $H$ at } H_{MaxID}$      
        \For{$pt \in cho_{cords}$} \label{alg: select_points1}
            \If{$pt.x, pt.y = p.x, p.y$} \Comment{if already chosen} 
                \State $H[H_{MaxID}, 0] \gets -\infty$
                \State $H_{MaxID} \gets \text{get maximum reward index in }H$ 
                \State $p.x, p.y \gets \text{x,y value at } H_{MaxID}$       
            \EndIf
        \EndFor \label{alg: select_points2}  
        \State $cho_{cords} \gets p$
        \State $goals \gets  p$
    \EndIf
    \State \textbf{return} $goals$
\EndFunction
\end{algorithmic}
\end{algorithm}

\begin{algorithm}
\small
\caption{Update Rewards}
\label{alg: update_rewards}
\begin{algorithmic}[1]
\Function{$upd\_rewards$}{$cho_{cords}$, $H$}
    \State $H_{MaxVal} \gets \text{max. reward in $H$}$  
    \State $K \gets \text{Equation \ref{eq: K}}$
    \For{$c \in cho_{cords}$ \& $g \in H$} \label{alg: update_rewards0}
        \If{$c_{x,y} = g_{x,y} $} \Comment{If goal already chosen then select reward to -infinity}
            \State $H[g] = -\infty$ \label{alg: update_rewards1}
        \EndIf
        \If{$H[g] \neq -\infty$}
            \Comment{compute distance from chosen goals}
            \State $d2 \gets \sqrt{(c_x - g_x)^2  + (c_y - g_y)^2}$ \label{alg: update_rewards2}
            \If{$d2 \neq 0$}                
                \State $H[g] \gets$ Equation \ref{eq: R_update}  \label{alg: update_rewards3}
            \Else
                \State $H[g] = -\infty$
            \EndIf
            \EndIf
    \EndFor
    \State \textbf{return} $H$
\EndFunction
\end{algorithmic}
\end{algorithm}

Algorithm \ref{alg: select_points} takes as input a list of points $uni_{pts}$ along with their rewards $uni_{ptsU}$ (computed with Equation 5), and updates the $H$ matrix using Algorithm \ref{alg: update_rewards} (line \ref{alg: select_points0}). It checks whether the frontier corresponds with the already selected goal, discarding the point in this case, and then updates the reward as described in Equations \ref{eq: K}-\ref{eq: R_update}. The goal corresponding to the higher reward for the current agent, updated with the described policies, is eventually passed back to the agent.

Algorithm \ref{alg: update_rewards} takes as input the chosen frontier points $cho_{cords}$ which are the previous assigned goal points from the previous iteration and $H$ matrix and starts by finds the maximum reward in $H$ and calculating a parameter $K$. It then iterates over each chosen frontier coordinate and every goal $g$ in $H$. If a goal’s coordinates match a chosen coordinate, its reward is set to $-\infty$, effectively removing it from consideration. For other goals, it calculates the Euclidean distance to the chosen coordinate and updates the reward using Equation \ref{eq: R_update} if the distance is non-zero. Finally, the updated reward matrix $H$ returned.

As described before, agents are managed asynchronously with a priority approach. The priority assigned to robots can lead to having one or more robots with low priority being stuck because they are always prioritized by higher-priority agents. To avoid this issue, the server also considers the number of requests unrelated to each agent. Once this number exceeds a certain predefined threshold \texttt{GOAL\_SKIP\_WAIT}, the corresponding agent will be associated with the higher priority. This approach prevents robots from getting stuck and distributes goals more uniformly.

\begin{figure}[H]
    \centering
    \includegraphics[width=14cm, height=7cm]{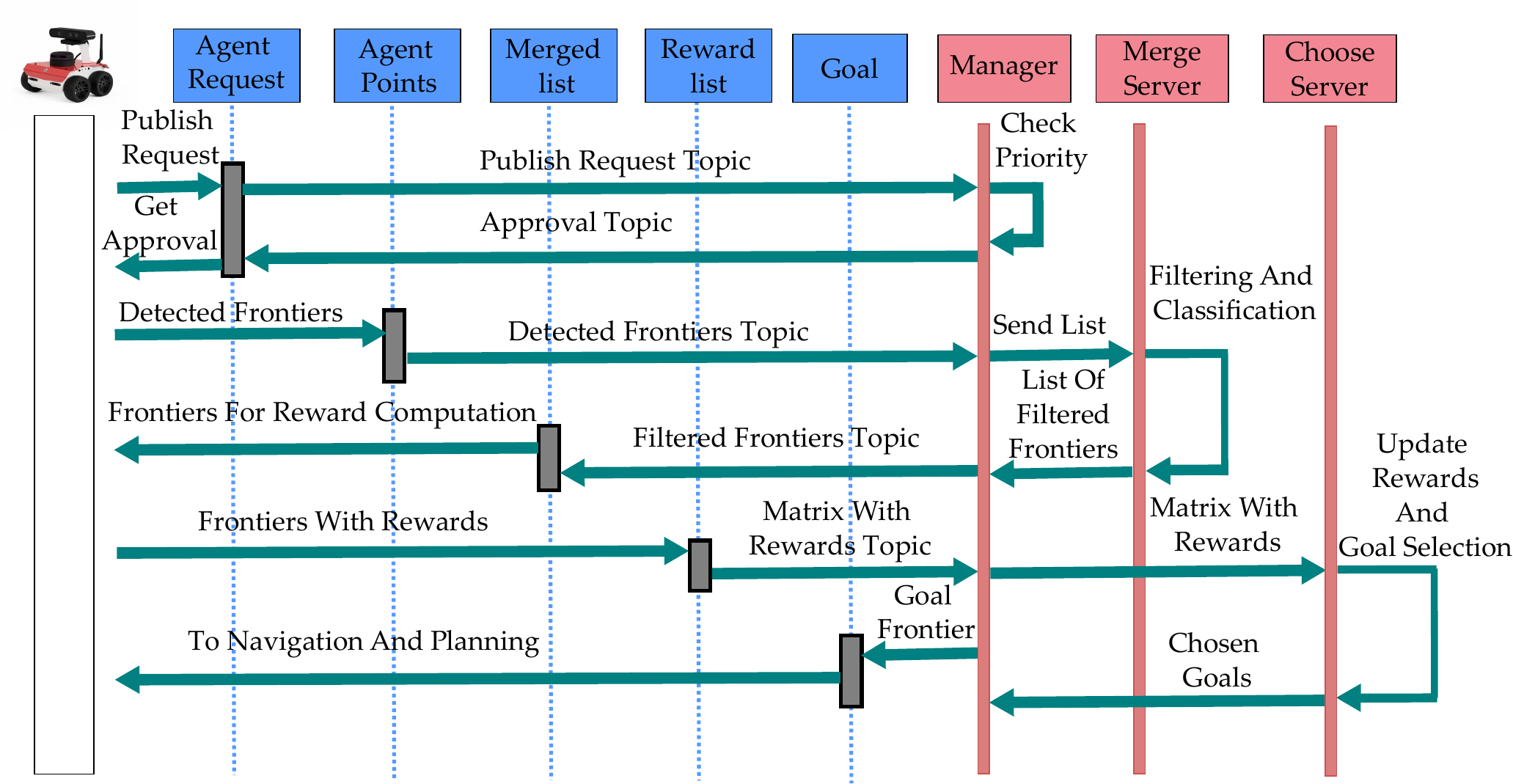}
    \caption{Sequence diagram. Local nodes of each robot (blue) and server (red).}
    \label{flow_chart}
\end{figure}

\subsection{Program Execution Flow}\label{ch4:Program Execution flow}

In this section, we provide a comprehensive explanation of the detailed execution flow involved in the goal frontier assignment process, a critical aspect of our approach. This process is represented in Figure \ref{flow_chart}, which offers a clear depiction of the various steps and decision points within the workflow. It is essential to emphasize that the methodology described here is designed to function without any explicit coordination or communication between the agents. In this context, each agent operates fully autonomously, making decisions independently based solely on its local information and reward computations. This lack of coordination implies that the agents do not share information or collaborate in the goal assignment process, thereby underscoring the decentralized nature of the approach. As a result, the decision-making process for each agent is entirely self-contained, with no reliance on the actions or states of the other agents within the system. The sequence of actions unfolds as follows:
\begin{enumerate}

    \item \textit{Publish Request}: Initially, the robot sends a request to the server by publishing a message on a dedicated topic. This publication process is uniform for all agents in the system, allowing the server to prioritize robot assignments effectively. 
    \item \textit{Approval and Execution}: Once the agent receives approval from the server, it proceeds with task execution. It gathers all the points it has detected and publishes them on a dedicated topic. The server, through subscription, collects the lists of points recently published by all agents. Subsequently, it employs Algorithm \ref{alg: list_creation} to eliminate duplicate points from the lists and Algorithm \ref{alg: is_near_border} to distinguish frontier points from others. The resultant list, once compiled, is then published on a designated topic. This list typically contains a set of points, constrained by parameters such as \texttt{MIN\_PTS} and \texttt{MAX\_PTS}.
    \item \textit{Reward Computation}: With the obtained list, the agent proceeds to calculate the reward according to the utility function in Equation \ref{eq:18}. 
    \item \textit{Server-side Reward Update}: Upon receiving the reward matrix from an agent, the server updates its records using Algorithm \ref{alg: update_rewards}. This process ensures that the server maintains an up-to-date assessment of the associated rewards.
    \item \textit{Goal Selection}: Finally, the server selects the goal that offers the highest reward from among the updated rewards as described in Algorithm \ref{alg: update_rewards}. This chosen goal becomes the target that the robot is assigned to reach and explore.
   

\end{enumerate}

This entire procedure, encompassing steps 1 through 5, is executed each time an agent reaches its previous goal.

\section{Experimental Evaluation}
\subsection{Simulation Environment}\label{Simulation results}

To evaluate the proposed approach, simulations were performed with the simulation environment Gazebo on a PC with an Intel Core i7\textsuperscript{\textregistered} (32GB RAM) and NVIDIA RTX 1000 GPU, equipped with Ubuntu 20.04 and ROS Noetic. The OG maps have a resolution of 0.1$m$/cell. Practically, a team of RosBots 2, equipped with lidar sensors were deployed in a modified version of the Willow Garage (W.G)\footnote[5]{\url{https://github.com/arpg/Gazebo/}} environment and that of HOS (from Section \ref{Related Work}), with an area of 2071 $m^2$ and 1243 $m^2$, respectively. 

We compared our proposed approach against 1) Frontier Detection based Exploration (Frontier) \cite{yamauchi} which uses a greedy frontier exploration strategy without any SLAM uncertainty quantification. 2) The method of \cite{NP4} by converting it into a multi-robot system namely MAGS. 

Since our proposed approach aims at environment exploration while working on minimum frontier points with efficient AC-SLAM, we decided to use the following performance metrics: 
\begin{itemize}
\item \textit{percentage of map coverage}, to quantify the evolution of the covered map concerning the ground truth map.
\item \textit{number of frontier points}, to measure the average points reduction (corresponding to a decreased computational cost) achieved with the method described in Section \ref{Filtering and classification}.

\item \textit{Map quality}, we compared metrics measuring Structural Similarity Index Measurement (SSIM) $\in [0,1]$, Root Mean Square Error (RMSE), and Alignment Error (AE) with reference to ground truth maps.

\end{itemize}
 We conducted 10 simulations of 20 minutes each for both W.G and HOS using Frontier, MAGS, and our method resulting in a total simulation time of 10 hours. \texttt{PER\_UNK}, \texttt{RAD}, \texttt{MIN\_PTS}, \texttt{MAX\_PTS} and \texttt{GOAL\_SKIP\_WAIT} were initialized to 60 \%, 1$m$, 0, 10 and 5 respectively.

\begin{figure}[H]
    \centering

    \includegraphics[width=\linewidth, height = 5.5cm]{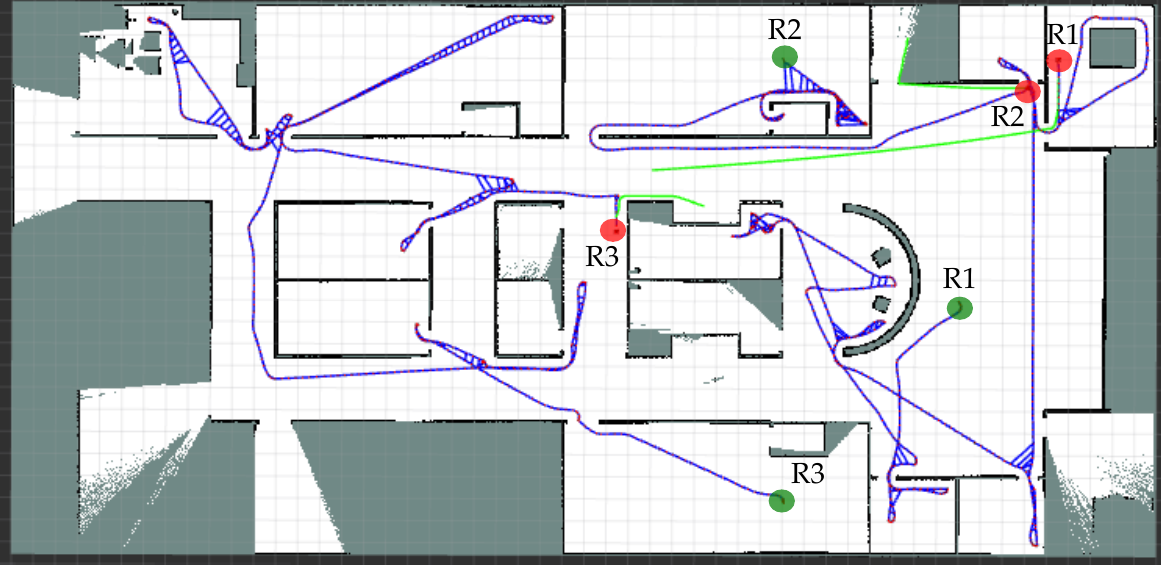}
    \caption*{(a) HOS environment OG map and pose graphs.}

    \begin{subfigure}[b]{0.32\linewidth}
        \centering
        \includegraphics[width=4.4cm,height=4.5cm]{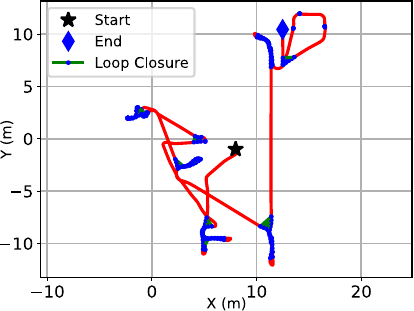}
        \caption{Robot 1 (R1)}
    \end{subfigure}
    \hfill
    \begin{subfigure}[b]{0.32\linewidth}
        \centering
        \includegraphics[width=4.4cm,height=4.5cm]{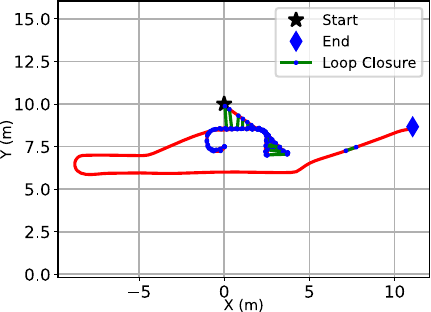}
        \caption{Robot 2 (R2)}
    \end{subfigure}
    \hfill
    \begin{subfigure}[b]{0.32\linewidth}
        \centering
        \includegraphics[width=4.4cm,height=4.5cm]{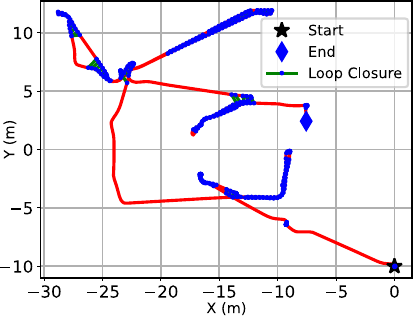}
        \caption{Robot 3 (R3)}
    \end{subfigure}

    \caption{Top: Final OG map using our method. Bottom: Individual pose graphs.}
    \label{fig:hospital_combined}
\end{figure}

\begin{figure}[H]
    \centering

    \includegraphics[width=\linewidth,height = 4.5cm]{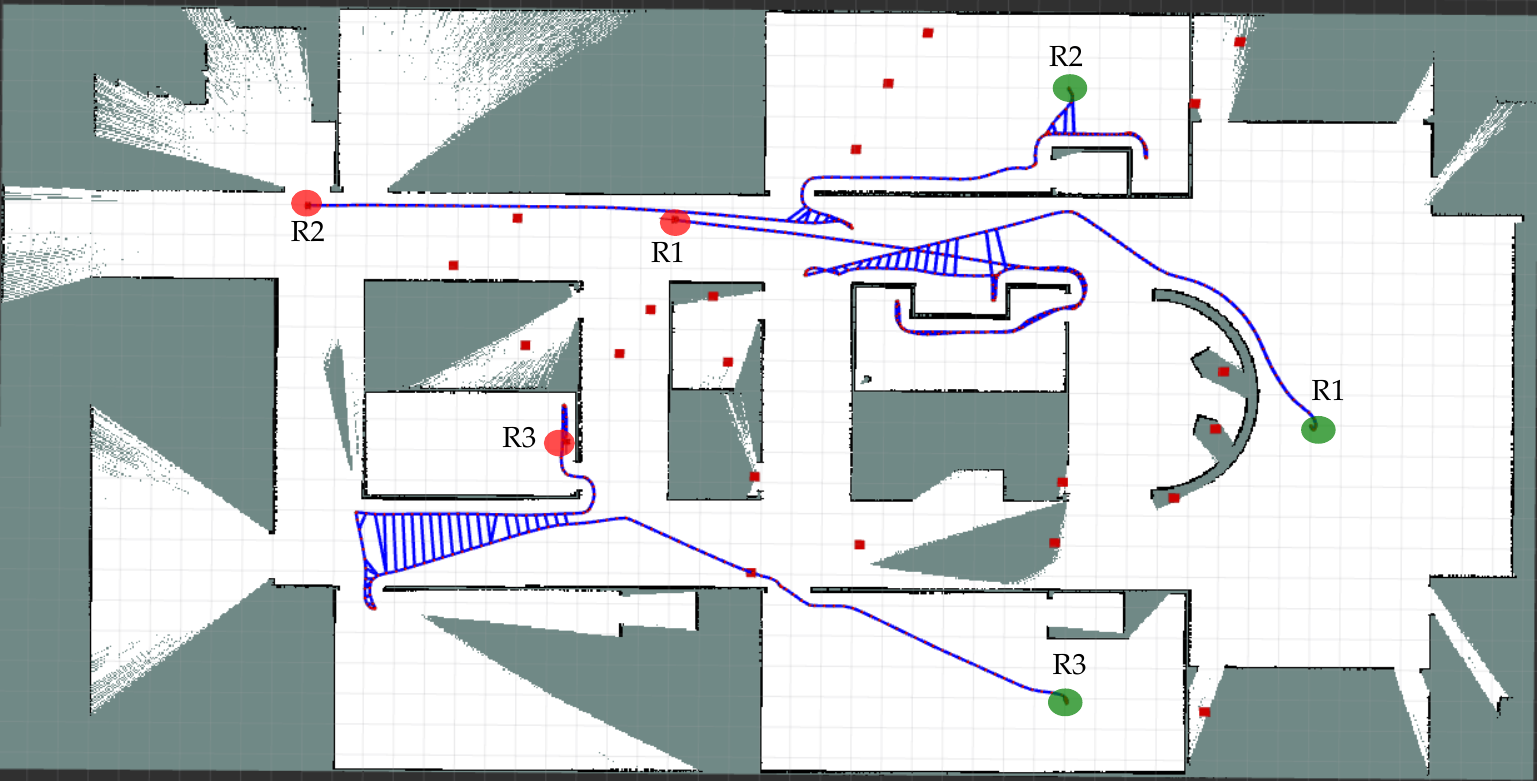}
    \caption*{(a) HOS environment OG map and pose graphs.}

    \begin{subfigure}[b]{0.32\linewidth}
        \centering
        \includegraphics[width=4.4cm,height=4.5cm]{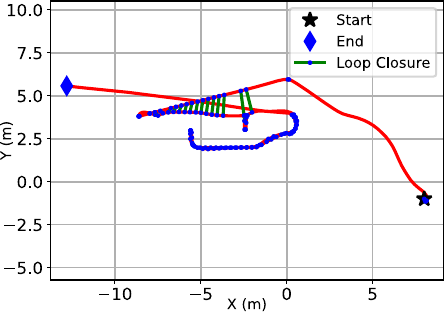}
        \caption{R1}
    \end{subfigure}
    \hfill
    \begin{subfigure}[b]{0.32\linewidth}
        \centering
        \includegraphics[width=4.4cm,height=4.5cm]{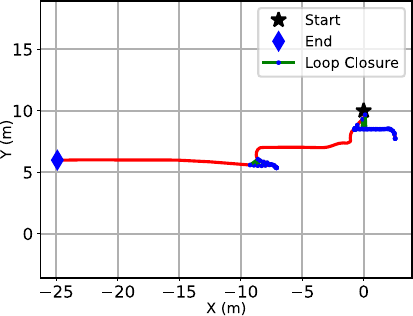}
        \caption{R2}
    \end{subfigure}
    \hfill
    \begin{subfigure}[b]{0.32\linewidth}
        \centering
        \includegraphics[width=4.4cm,height=4.5cm]{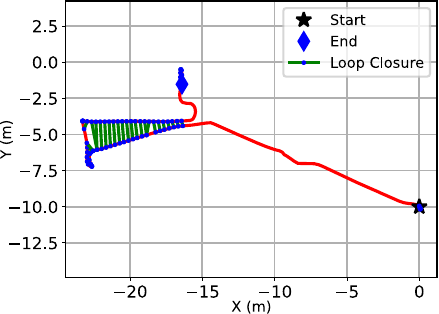}
        \caption{R3}
    \end{subfigure}

    \caption{Top: Final OG map using MAGS method. Bottom: Individual pose graphs.}
    \label{fig:hospital_combined_mags}
\end{figure}

\begin{figure}[H]
    \centering

    \includegraphics[width=\linewidth,height = 4.5cm]{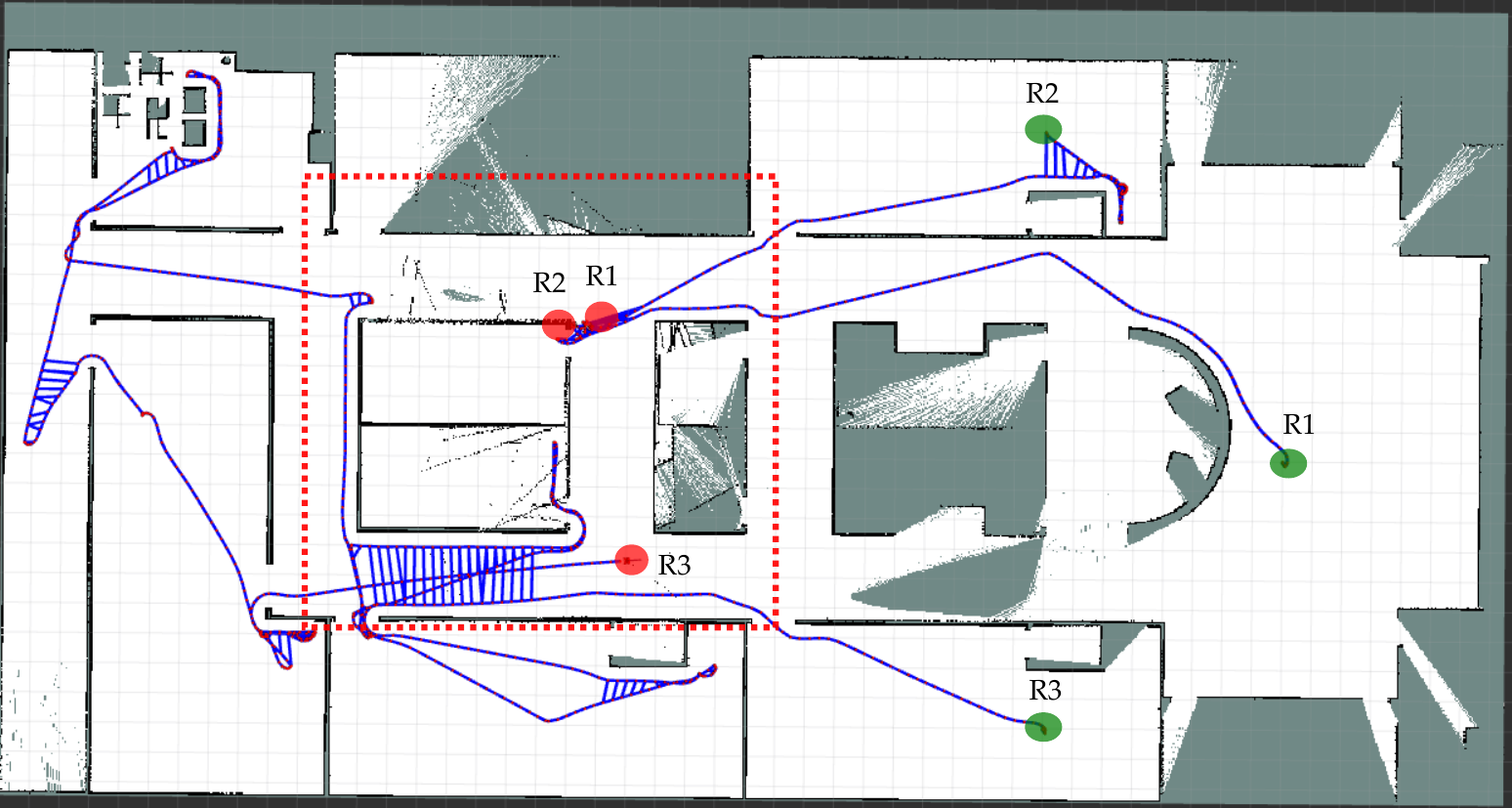}
    \caption*{(a) HOS environment OG map and pose graphs.}

    \begin{subfigure}[b]{0.32\linewidth}
        \centering
        \includegraphics[width=4.4cm,height=4.5cm]{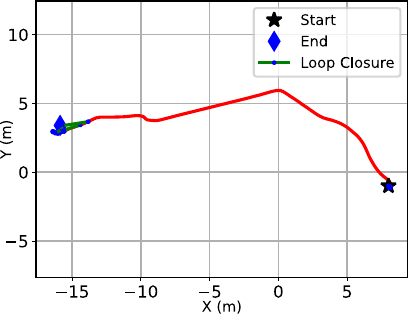}
        \caption{R1}
    \end{subfigure}
    \hfill
    \begin{subfigure}[b]{0.32\linewidth}
        \centering
        \includegraphics[width=4.4cm,height=4.5cm]{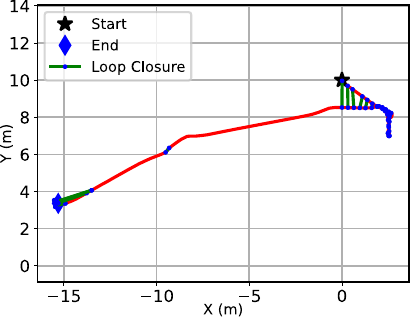}
        \caption{R2}
    \end{subfigure}
    \hfill
    \begin{subfigure}[b]{0.32\linewidth}
        \centering
        \includegraphics[width=4.4cm,height=4.5cm]{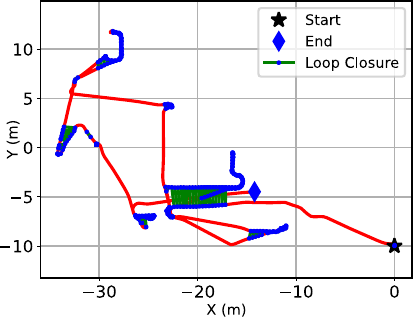}
        \caption{R3}
    \end{subfigure}

    \caption{Top: Final OG map using Frontier method. Bottom: Individual pose graphs.}
    \label{fig:hospital_combined_frontier}
\end{figure}

Figures \ref{fig:hospital_combined}, \ref{fig:hospital_combined_mags}, and \ref{fig:hospital_combined_frontier} show the resulting OG maps and pose graphs generated using Open Karto SLAM, including loop closures, for the AWS Hospital environment. These maps correspond to our approach, MAGS, and Frontier, respectively. The Figures also highlight the starting positions (in green) and final positions (in red) of the three agents after 30 minutes of exploration.

From Figure \ref{fig:hospital_combined}, we observe that our method achieves accurate mapping with 80\% coverage while maintaining good SLAM accuracy, as the agents actively seek loop closures. In Figure \ref{fig:hospital_combined_mags}, we also observe good localization and mapping accuracy, though the coverage is lower at 60\%, and fewer loop closures are performed. Finally, in Figure \ref{fig:hospital_combined_frontier}, the agents explore 70\% of the area—more than MAGS (Figure \ref{fig:hospital_combined_mags})—but with reduced SLAM accuracy. This results in poor localization and mapping performance, as illustrated by the dotted rectangle where agents R1 and R2 fail to accurately map the environment due to limited localization accuracy. 

\begin{figure}[H]
\captionsetup[subfigure]{justification=centering}
     \centering
     \begin{subfigure}[b]{0.4\textwidth}
         \centering
         \includegraphics[width=6.3cm,height=5cm]{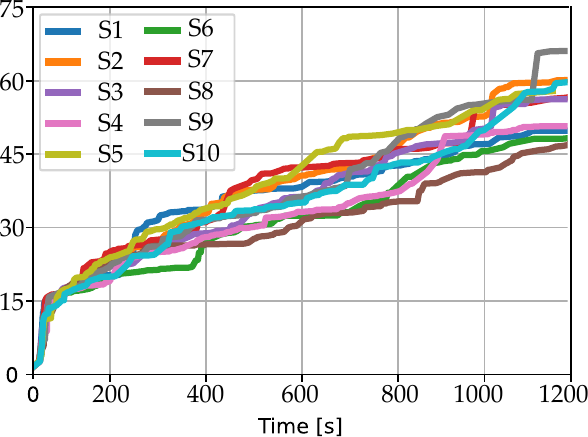}
         \caption{W.G.}
         \label{fig:Multiple simulations on Willow Garage Environment1}
     \end{subfigure}
     \hfill
     \begin{subfigure}[b]{0.45\textwidth}
         \centering
         \includegraphics[width=6.3cm,height=5cm]{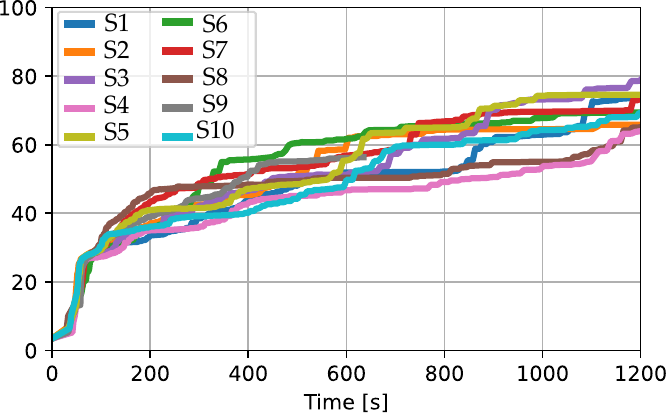}
         \caption{HOS.}
         \label{fig:Multiple simulations on Willow Garage Environment2}
     \end{subfigure}
    
     \hfill
        \caption{Evolution of \% area coverage}
        \label{fig:Multiple simulations on Willow Garage Environment}
\end{figure}

Figure \ref{fig:Multiple simulations on Willow Garage Environment1} shows the evolution of the percentage [\%] of map area discovered in W.G and HOS environments in 10 simulations using our approach and 3 robots. In both environments, the agents initially discover 15\% and 20\% area and eventually manage to cover 48\% (994$m^2$) and 70\% (870$m^2$) area respectively.

\begin{figure}[H]
    \centering
    \includegraphics[width=13cm, height =5cm]{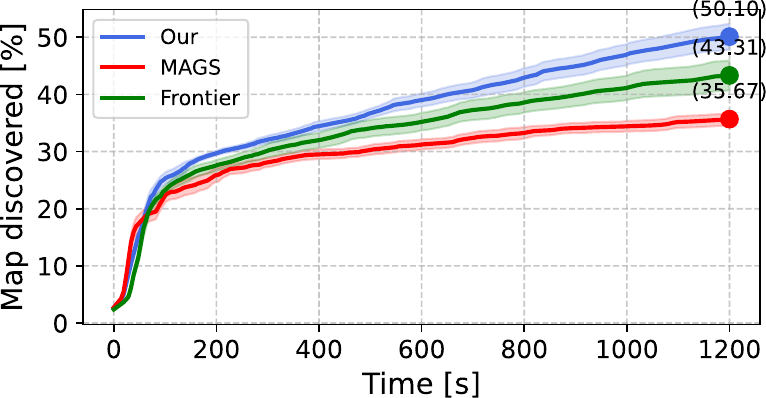}
\caption{The average percentage along with the standard deviation of the explored area in the W.G environment.}

    \label{fig: w.g_3robots}
\end{figure}

\begin{figure}[H]
    \centering
    \includegraphics[width=13cm, height =5cm]{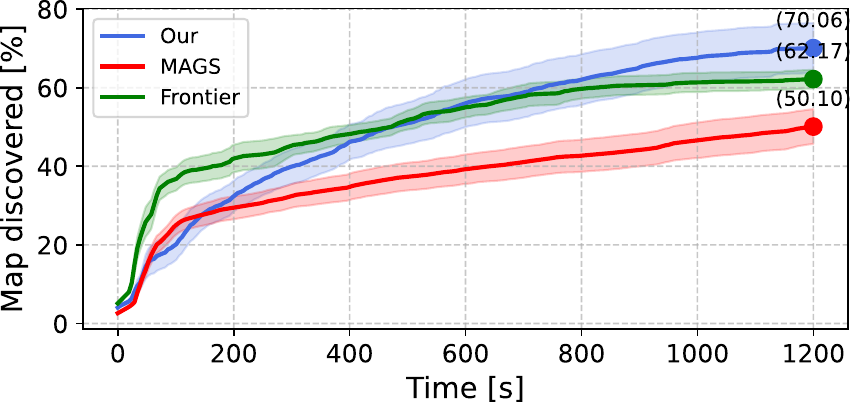}
\caption{The average percentage along with the standard deviation of the explored area in the HOS environment.}

    \label{fig: hos_3robots}
\end{figure}

Figures \ref{fig: w.g_3robots} and \ref{fig: hos_3robots} show the average percentage of the map explored by three robots in the W.G. and HOS environments, respectively, using our approach (blue), MAGS (red), and Frontier (green). In Figure \ref{fig: w.g_3robots}, our approach explores an area of 1040$m^2$, which is 43\% and 16\% more than the areas covered by MAGS and Frontier, respectively. Similarly, in Figure \ref{fig: hos_3robots}, our method covers 870$m^2$, representing 40\% and 12\% more area than MAGS and Frontier, respectively.

\begin{figure}[H]
\centering
  \subfloat[W.G. \label{ch4:fig:bargraph1}]{%
       \includegraphics[width=6.3cm,height=4cm]{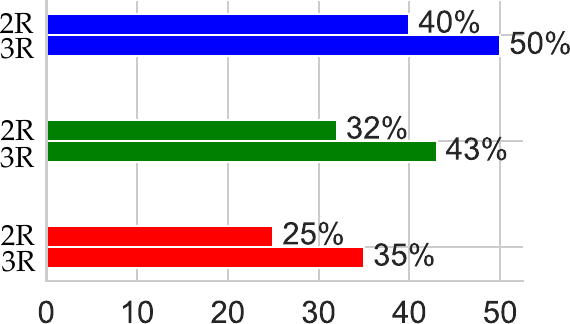}}
    \hfill
  \subfloat[HOS. \label{ch4:fig:bargraph2}]{%
        \includegraphics[width=6.3cm,height=4cm]{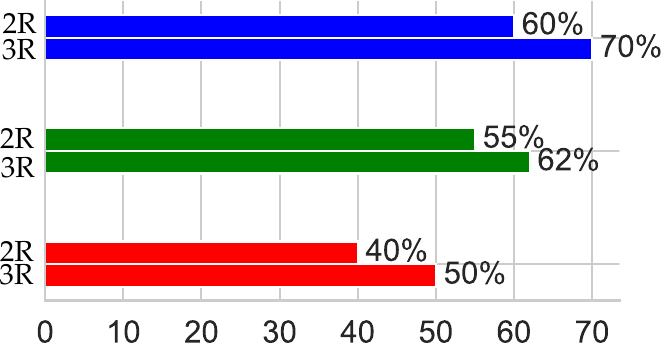}}        
  \caption{Average percentage of explored map in both simulation environments.}
  \label{ch4:fig:bargraph} 
\end{figure}

Additionally, we observe that in both environments, the Frontier method outperforms MAGS in terms of exploration area. This is because Frontier follows a greedy exploration strategy without incorporating uncertainty quantification, leading to more extensive exploration but at the expense of SLAM map quality and localization. Figure \ref{ch4:fig:bargraph} plots the average percentage of maps discovered using two and three agents (subscript $X$R denotes the number of robots). It can be deduced that our approach outperforms other approaches considerably resulting in more average area coverage in both environments. 

\begin{figure}
\centering
  \subfloat[W.G. \label{fig:points detected1}]{%
       \includegraphics[width=6.5cm,height=4cm]{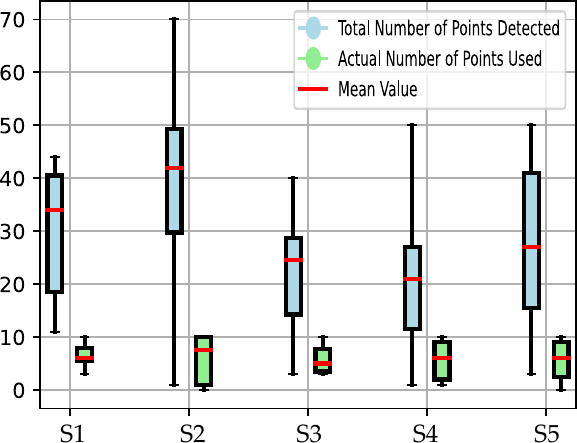}}
    \hfill
  \subfloat[HOS. \label{fig:points detected2}]{%
        \includegraphics[width=6.5cm,height=4cm]{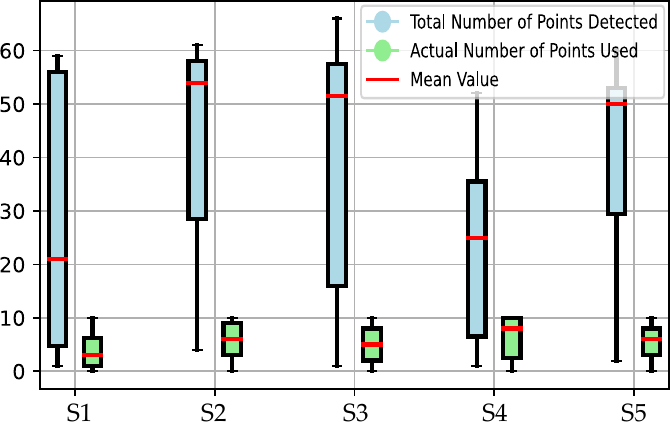}}        
  \caption{Total points detected and actual points used in both environments.}
  \label{fig:points detected} 
\end{figure}

Regarding computational complexity and reducing the number of processed frontier points, Figure \ref{fig:points detected1} provides an insight into how the number of points used is reduced in the W.G environment. We can deduce that using the methods in Section \ref{Filtering and classification} and setting \texttt{PER\_UNK} $ = 60\%$, \texttt{RAD}$ = 1m$ we manage to drastically reduce the average number of points processed from 35 to 8 for S1 (77\%), 41 to 9 for S2 (78\%), 27 to 7 for S3 (74\%), 20 to 8 for S4 (60\%) and 28 to 8 for S5 (71\%) respectively, also exploring the environment more efficiently. This reduction in points is directly related to the computational complexity, as fewer points require computing the utility function with lesser frequency, hence accelerating the overall performance of the proposed system.

The same comparison has been carried out in the smaller HOS environment, in Figure \ref{fig:points detected2} the usage of the \textit{Filtering and Classification} method leads to a meaningful reduction in the number of points processed from 21 to 4 for S1 (80\%), 55 to 6 for S2 (89\%), 51 to 5 for S3 (90\%), 25 to 9 for S4 (64\%) and 50 to 7 for S5 (86\%) respectively. The average values are higher compared to those in Figure \ref{fig:points detected1} since also the environment has more obstacles than W.G.

Table \ref{tab: PER_UNK} presents the usage of \texttt{PER\_UNK}, which represents the percentage of unknown cells considered within a given radius when computing the information gain of a frontier candidate. The table also includes \texttt{RAD}, showing how the radius values change when the list of points is recomputed using three robots with our approach.

We observe that in W.G, \texttt{PER\_UNK} decreases to $\leq 40\%$ (from an initial threshold of $60\%$), indicating the re-computation of the list. This occurs because W.G has fewer obstacles than HOS, leading to increased computational effort to maximize $uni_{pts}$, as explained in Algorithm \ref{alg: points_list_loop}.

In contrast, for HOS, \texttt{PER\_UNK} remains at $\leq 60\%$, suggesting fewer list re-computations on the server, thereby reducing computational cost. Additionally, in both environments, \texttt{RAD} remains predominantly at $1m$, indicating that the list size remains $\leq$ \texttt{MAX\_PTS}, which reflects lower computational effort required by Algorithm \ref{alg: points_list_loop} to reduce the number of points. \\

\begin{table}[h]
\centering
\begin{tabular}{|c|c|c|c|c|c|c|}
\hline
\multirow{2}{*}{\textbf{Env.}} &   \multicolumn{3}{c|}{\textbf{\texttt{PER\_UNK} [\%]}} & \multicolumn{3}{c|}{\textbf{\texttt{RAD} [$m$]}} \\

& 60 & 50 & $\leq$ 40 & 1.00 & 1.25 & $\leq$ 1.50  \\
\hline
\multirow{1}{*}{W.G} & 34.5 & 1.4 & \textbf{64.0} & \textbf{87.0} & 1.8 & 9.7 \\
\hline
\multirow{1}{*}{HOS} & \textbf{67.3} & 4.3 & 28.2  & \textbf{76.2} & 5.1 & 8.5 \\
\hline

\end{tabular}
\caption{Impact of \texttt{PER\_UNK} threshold and \texttt{RAD} values on frontier list computation in both environments.}
\label{tab: PER_UNK}
\end{table}

\begin{table}[h]
\centering
\begin{tabular}{|c|c|c|c|c|}
\hline
\textbf{Env.} & \textbf{Method} & \textbf{SSIM} & \textbf{RMSE} & \textbf{AE} \\
\hline        
\multirow{3}{*}{W.G}
  & Our      & 0.80           & 4.53          & 21.68 \\
  & MAGS     & \textbf{0.86}  & 6.34          & 28.39 \\
  & Frontier & \textit{0.20}  & \textbf{10.04} & \textbf{40.89} \\
\hline
\multirow{3}{*}{HOS}
  & Our      & \textbf{0.78}  & 4.02          & 21.39 \\
  & MAGS     & 0.72           & 6.39          & 29.98 \\
  & Frontier & 0.35           & \textbf{12.67} & \textbf{42.89} \\
\hline
\end{tabular}		
\caption{Comparison of map quality metrics (SSIM, RMSE, AE) for both environments.}
\label{tb:accuracy_comparison2}
\end{table}

Regarding the visual analysis of the maps and the evaluation of map quality metrics, the results, on average, appear promising, as shown in Table \ref{tb:accuracy_comparison2} using three robots. In almost all cases, our method yielded lower RMSE and AE while achieving higher SSIM compared to the MAGS and Frontier methods. Furthermore, we observe that the Frontier method, in comparison to MAGS, explores the environment more extensively, as illustrated in Figure \ref{fig: w.g_3robots}, but at the cost of higher RMSE and AE, as well as lower SSIM.

The aforementioned simulation results highlight the efficiency of our approach compared to state-of-the-art methods. To further validate our methodology, we conducted experiments with a team of ground robots in a real-world environment. The results of these experiments are presented in the following Section.

\subsection{Real Environment}
\label{Experimental results}
\begin{figure}[H]
\centering
  \subfloat[ROSbot 2R. \label{fig:Robot and experimental environment used1}]{%
       \includegraphics[width=3.5cm,height=3cm]{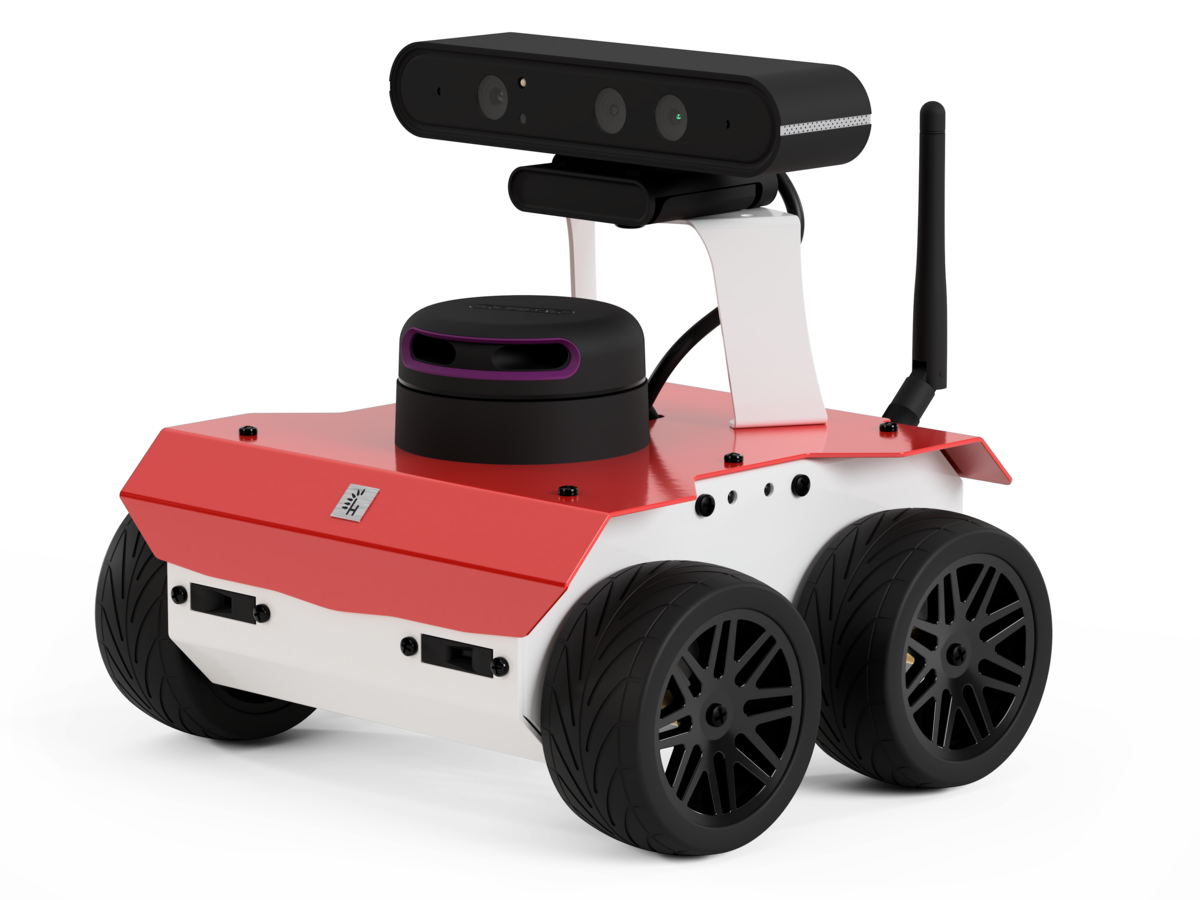}}
    \hfill
  \subfloat[Experimental environment. \label{fig:Robot and experimental environment used2}]{%
        \includegraphics[width=7cm,height=3.5cm]{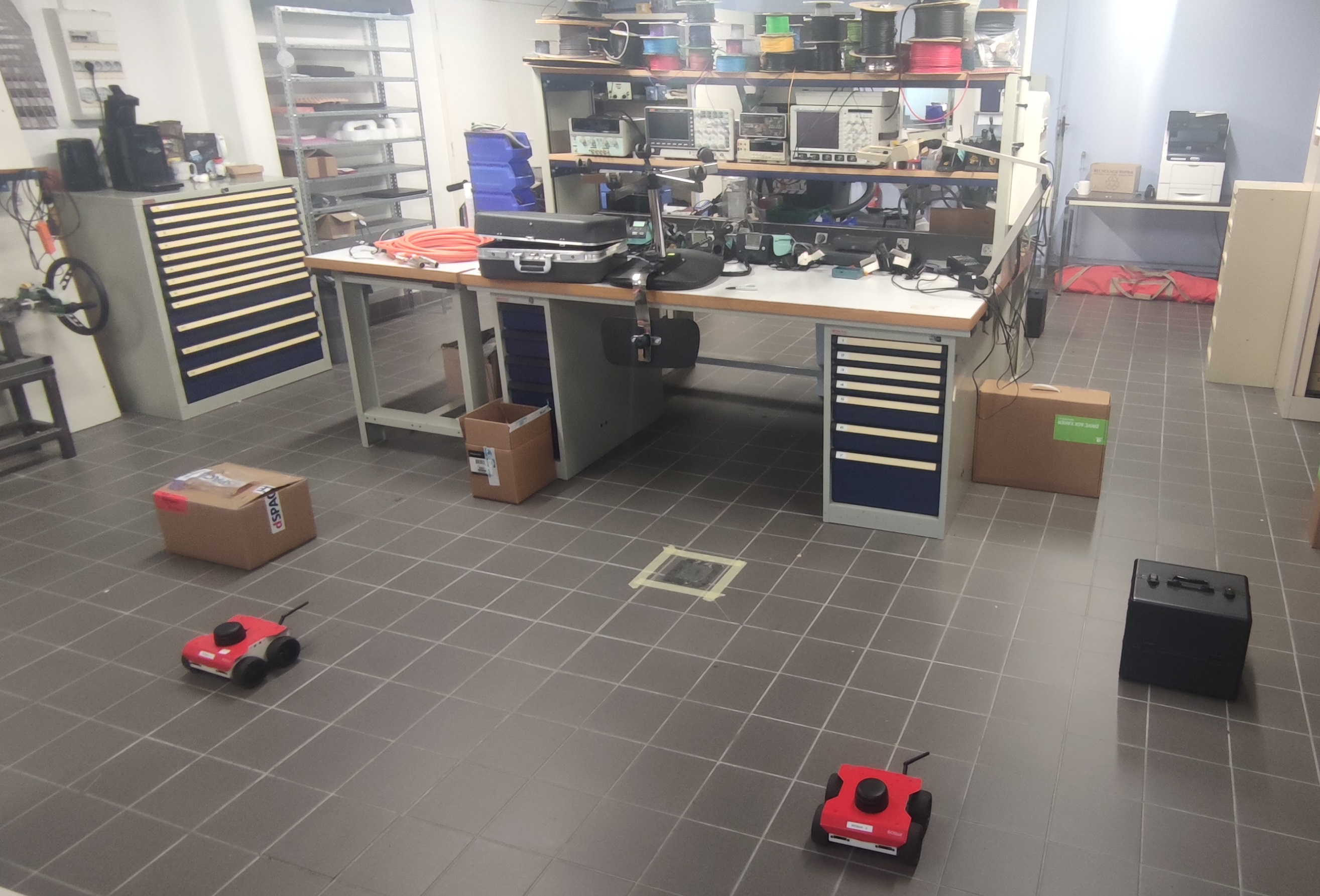}}      
        \caption{Robot and experimental environment used.}
        \label{fig:Robot and experimental environment used}
\end{figure}

\begin{figure}[H]
    \centering
    \subfloat[MAGS Experiment 1.\label{ch4:fig:explored area comparison1}]{\includegraphics[width=6.2cm, height=3.5cm]{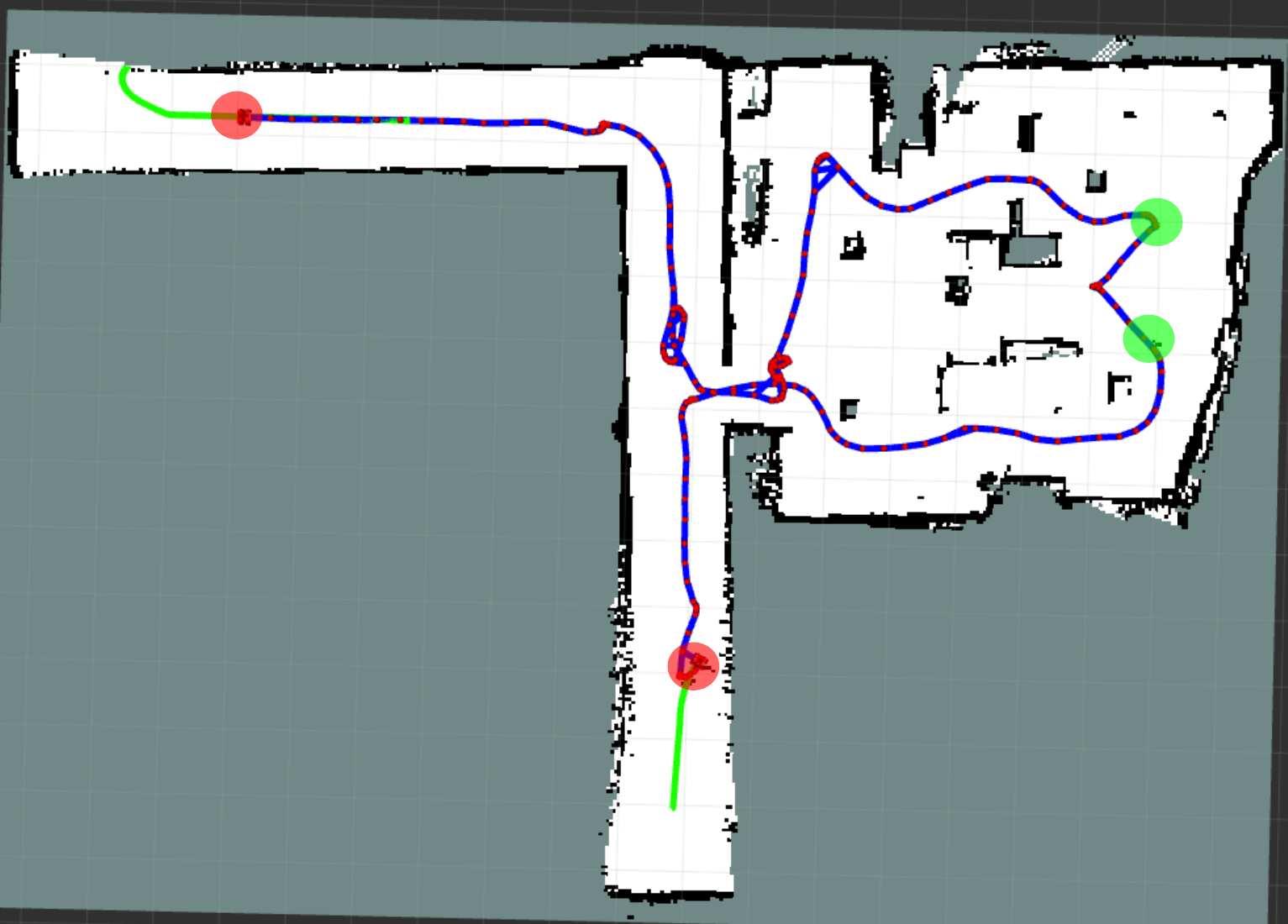}}
    \hfill
    \subfloat[MAGS Experiment 2.\label{ch4:fig:explored area comparison2}]{\includegraphics[width=6.2cm, height=3.5cm]{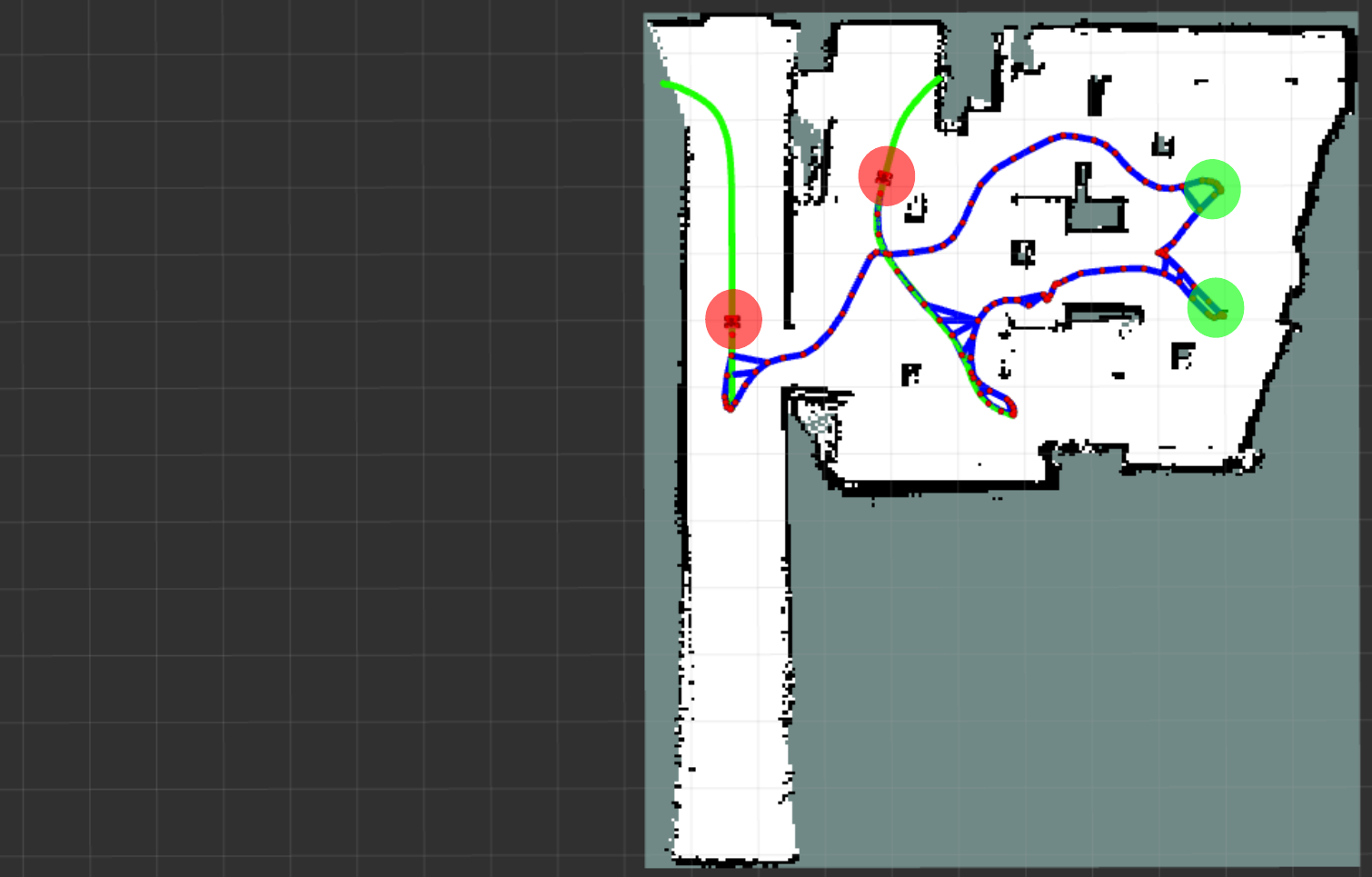}} \\
    \subfloat[Our method Experiment 1. \label{ch4:fig:explored area comparison3}]{\includegraphics[width=6.2cm, height=3.5cm]{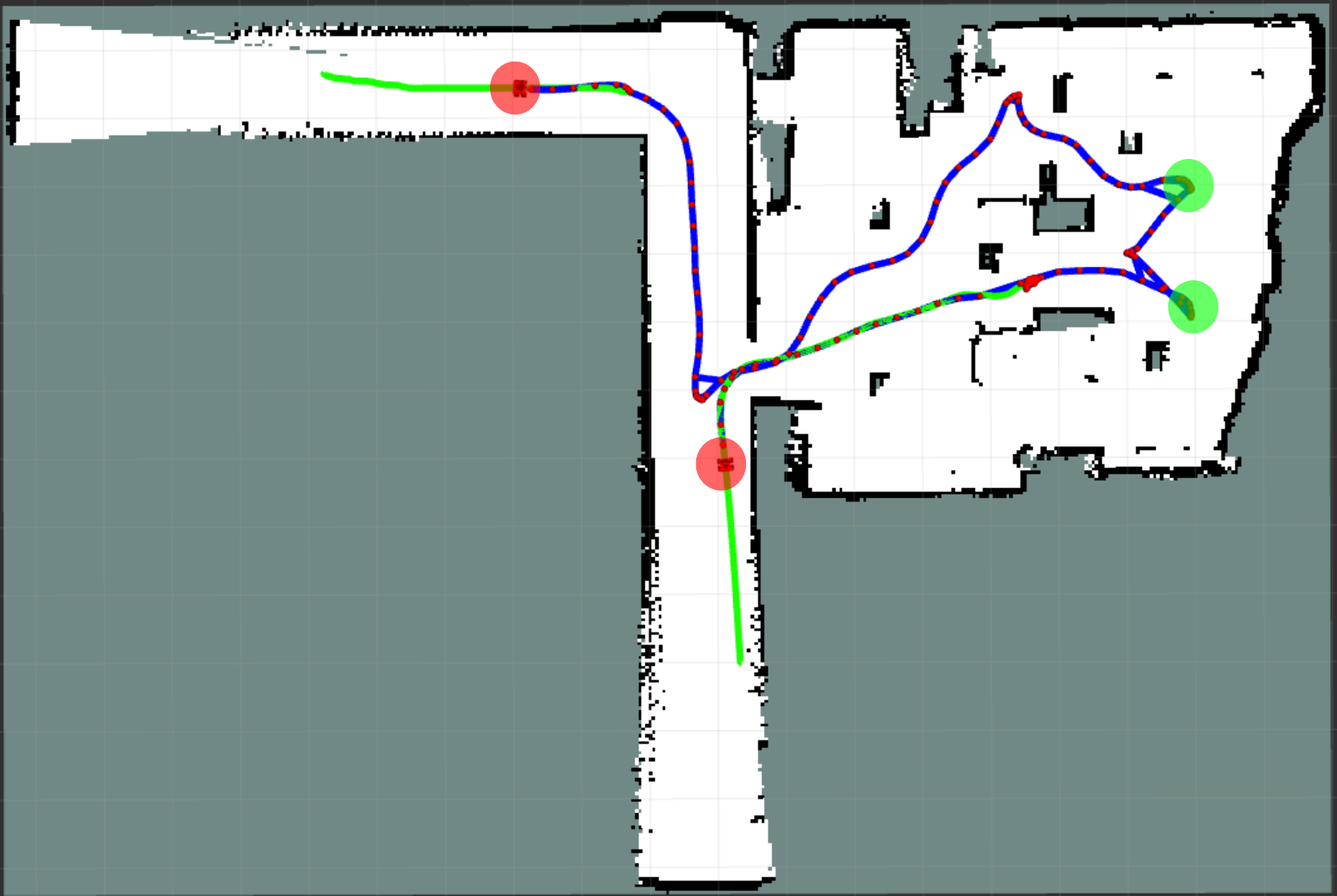}}
    \hfill
    \subfloat[Our method Experiment 2. \label{ch4:fig:explored area comparison4}]{\includegraphics[width=6.2cm, height=3.5cm]{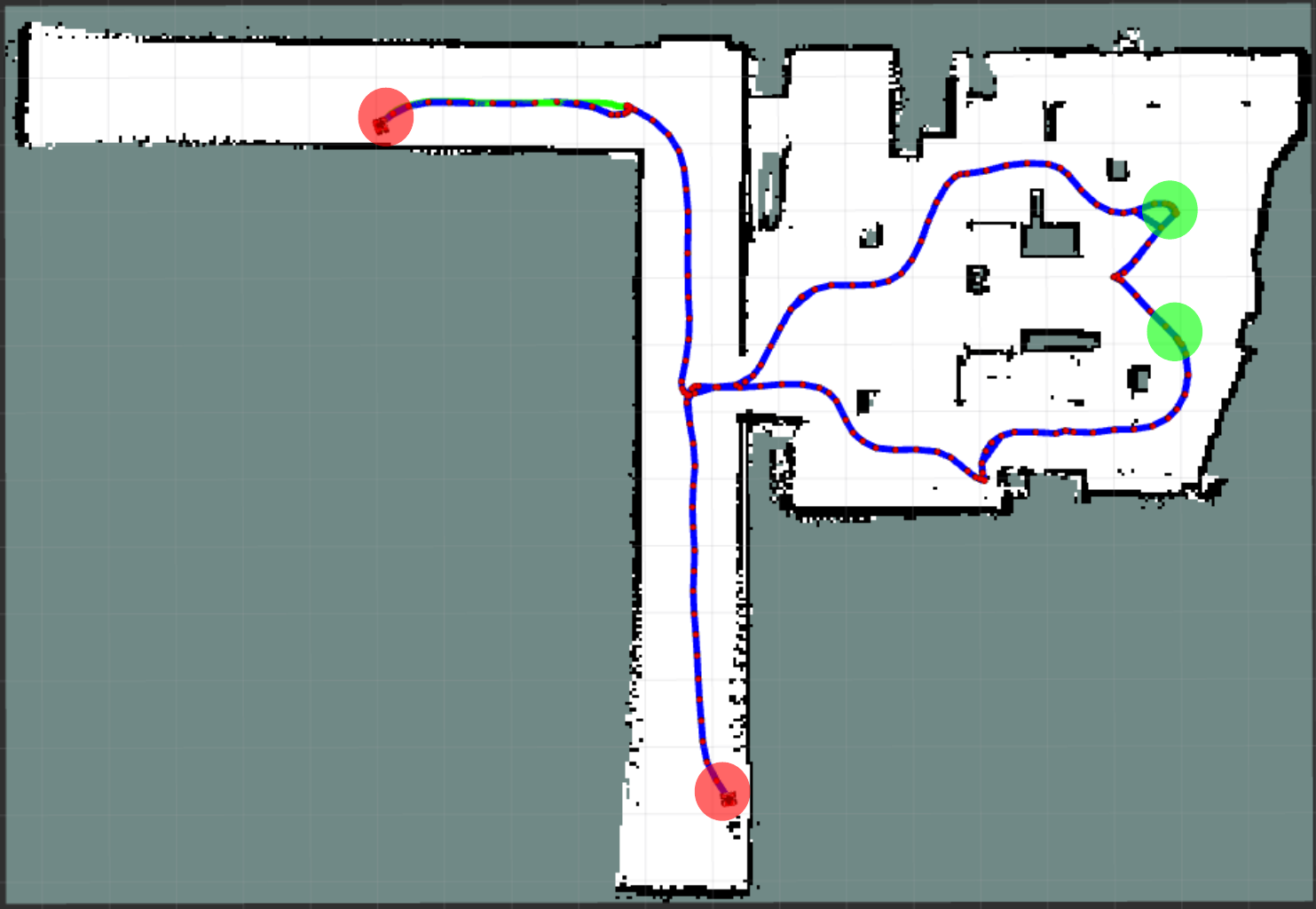}}    
      \caption{Comparison of explored area in MAGS and our methods using two robots.}
       \label{ch4:fig:explored area comparison} 
\end{figure}

Experiments in a real environment are performed using two ROSBot 2R robots\footnote[6]{\url{https://husarion.com/manuals/rosbot/}} with RPLidar A2 (Figure \ref{fig:Robot and experimental environment used1}) with ROS on Ubuntu 20.04.6 (LTS). The robots are equipped with an Intel Xeon\textsuperscript{\textregistered} W-2235 CPU (3.80GHz x 12), with 64Gb RAM and Nvidia Quadro RTX 4000 GPU. The environment consists of a room and two corridors measuring 81$m^2$ in total as shown in Figure \ref{fig:Robot and experimental environment used2}. Figure \ref{ch4:fig:explored area comparison} presents the computed OG map and SLAM pose graph, illustrating the start (green) and final (red) positions of both robots across four experiments using the MAGS method and our proposed approach.

In the MAGS method (Figures \ref{ch4:fig:explored area comparison1} and \ref{ch4:fig:explored area comparison2}), the robots frequently revisit previously explored areas, leading to excessive loop closures. This behavior reduces exploration efficiency and limits overall environment coverage. In contrast, our approach (Figures \ref{ch4:fig:explored area comparison3} and \ref{ch4:fig:explored area comparison4}) minimizes loop closures, allowing the robots to focus on unexplored areas and achieve broader coverage. This further demonstrates that while both methods produce accurate maps, the proposed approach enhances SLAM performance by prioritizing efficient exploration.

\begin{figure}[H]
\captionsetup[subfigure]{justification=centering}
     \centering
     \begin{subfigure}[b]{0.6\textwidth}
         \centering
         \includegraphics[width=8cm,height=5.5cm]{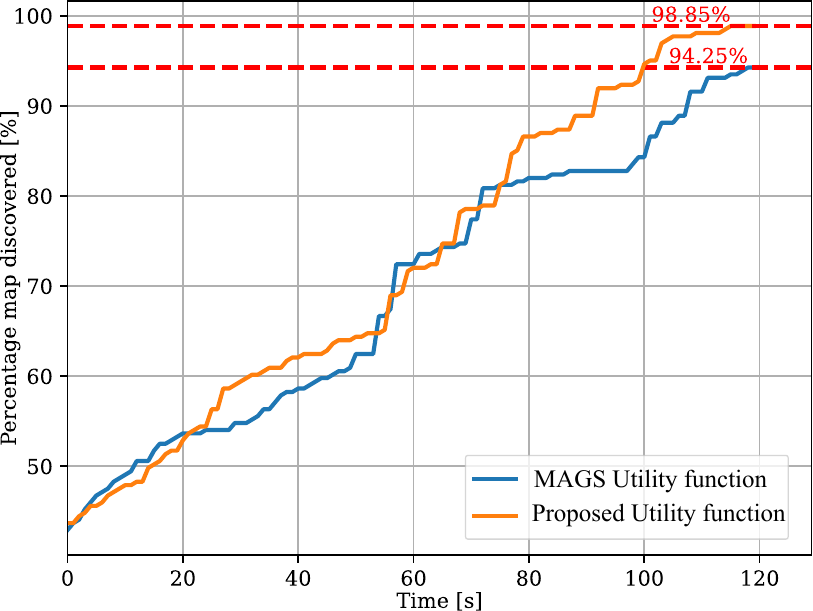}
         \caption{Comparison of mean values.}
         \label{fig:Comparision and points1}
     \end{subfigure}
     \hfill
     \begin{subfigure}[b]{0.36\textwidth}
         \centering         
         \includegraphics[width=5.6cm,height=3.7cm]{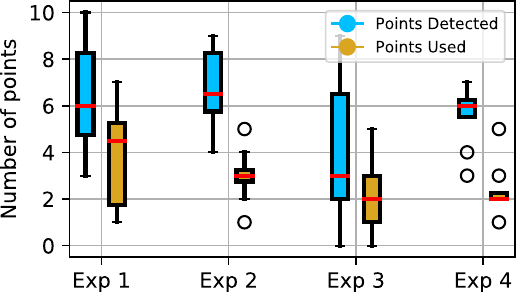}
         \caption{Points filtering.}
         \label{fig:Comparision and points2}
     \end{subfigure}
    
     \hfill
        \caption{(\ref{fig:Comparision and points1}) Comparison of mean values of map discovered with and without the proposed utility function. (\ref{fig:Comparision and points2}) Total points detected vs the actual points used.} 
        \label{fig:Comparision and points}
\end{figure}

Figure \ref{fig:Comparision and points1} shows the percentage of maps discovered over time. As for simulation results, we compared the proposed utility function with the 'MAGS' in four experiments, considering an exploration time of 2 minutes. It can be shown that the proposed approach achieves a  coverage of 98.85\%, higher than the coverage percentage achieved with MAGS (94.25\%). This implies a higher portion of map covered with the proposed approach (4.6\%) and hence proves the effectiveness of the proposed method. 

Figure \ref{fig:Comparision and points2} compares the number of points processed and detected across the four experiments. A significant reduction in the number of used points is observed: from 6 to 5 in Exp 1 (17\%), from 7 to 3 in Exp 2 (58\%), from 3 to 2 in Exp 3 (34\%), and from 6 to 2 in Exp 4 (67\%), leading to a notable decrease in computational load.

Due to the smaller size of the experimental environment, the exploration time was limited to just 2 minutes. Nevertheless, the increase in map coverage (4.6\%) was comparable to that achieved in the simulation (3\%–11\%). Additionally, the maximum number of detected and processed points was lower than in the simulation, with an average reduction of 44\% in the number of frontier points used.

\section{Conclusion}\label{Conclusion}

In this article, we presented a multi-robot collaborative active SLAM framework for efficient environment exploration. The proposed framework implements a utility function that incorporates the SLAM uncertainty and path entropy for the efficient selection of goal frontier candidates. We also proposed an efficient frontier filtering method that encourages sparsity while working on a reduced number of frontier candidates, hence providing a less computationally expensive solution. The overarching goal of our proposed system is to minimize map entropy, reduce SLAM uncertainty, and maximize the efficiency and coverage of the exploration process. By doing so, our system ensures that the robots generate accurate maps with low SLAM uncertainty while exploring unknown environments in the most efficient manner possible. The implementation of our framework leverages a ROS based client-server paradigm, which is integrated into a modular software architecture. This modularity facilitates ease of development, testing, and deployment, allowing each module to be independently developed and seamlessly integrated into the overall system. A thorough and detailed explanation of each module, along with the associated algorithms, is presented and discussed.

Extensive simulations on publicly available environments, along with real-world experiments, demonstrate the effectiveness of our approach. Compared to state-of-the-art methods, our framework achieves superior exploration efficiency, higher mapping accuracy, and a substantial reduction in computational costs. The results show that our method accelerates exploration while maintaining high-quality maps with reduced SLAM uncertainty, confirming its robustness and applicability in real-world scenarios.

Looking ahead, we plan to extend our approach to encompass visual active collaborative SLAM using heterogeneous robots. This extension will allow us to exploit visual features and leverage the unique perspectives provided by different types of robots, thereby further enhancing the robustness and applicability of our framework in diverse and dynamic environments.

\section*{Acknowledgement}
This work was carried out in the framework of the NExT Senior Talent Chair DeepCoSLAM, which was funded by the French Government, through the program Investments for the Future managed by the National Agency for Research ANR-16-IDEX-0007, and with the support of Région Pays de la Loire and Nantes Métropole.
This research was also supported by the DIONISO project (progetto SCN\_00320-INVITALIA), which is funded by the Italian Government.

\bibliography{sn-bibliography}

\section*{Statements and Declarations}

\begin{itemize}
\item Funding: This work was carried out in the framework of the NExT Senior Talent Chair DeepCoSLAM, which was funded by the French Government, through the program Investments for the Future managed by the National Agency for Research ANR-16-IDEX-0007, and  has been supported by the DIONISO project (progetto SCN\_00320 - INVITALIA) which is funded by the Italian Government.

\item Competing interests: The authors declare no conflict of interest. The funders had no role in the design of the study; in the collection, analyses, or interpretation of data; in the writing of the manuscript, or in the decision to publish the results.

\item Ethics approval and consent to participate: Not applicable.

\item Consent for publication: Not applicable.

\item Author contribution: Conceptualization, Muhammad Farhan Ahmed (M.F.A.), Matteo Maragliano (M.M.), Vincent Frémont (V.F.), Carmine Tomasso Recchiuto (C.T.R.); methodology, M.F.A. and M.M.; validation, V.F. C.T.R.; formal analysis, M.F.A.; investigation, M.M.; resources, V.F. and C.T.R.; data curation, M.F.A. and M.M.; writing original draft preparation, M.F.A. and M.M.; writing review and editing, C.T.R. and V.F.; visualization, M.F.A. and M.M.; supervision, V.F and C.T.R.; project administration, V.F., C.T.R; and funding acquisition, V.F. All authors have read and agreed to the published version of the manuscript.

\end{itemize}

\noindent

\end{document}